\documentclass[a4paper,fleqn]{cas-dc}

\usepackage[numbers]{natbib}

\usepackage{algorithm}
\usepackage{algpseudocode}
\usepackage{subfigure}
\usepackage{multirow,multicol}
\usepackage{hyperref}
\hypersetup{
    citecolor=green, 
    linkcolor=red, 
    urlcolor=green  
}
\usepackage{wrapfig}
\usepackage{flushend}

\def\tsc#1{\csdef{#1}{\textsc{\lowercase{#1}}\xspace}}
\tsc{WGM}
\tsc{QE}
\tsc{EP}
\tsc{PMS}
\tsc{BEC}
\tsc{DE}

\begin{document}
\let\WriteBookmarks\relax
\def\floatpagepagefraction{1}
\def\textpagefraction{.001}
\shorttitle{DCCMVC}
\shortauthors{Bo Li et~al.}

\title [mode = title]{Dual Consistent Constraint via Disentangled Consistency and Complementarity for Multi-view Clustering}                      
\tnotemark[1]

\tnotetext[1]{This work was supported by the National Natural Science Foundation of China under Grants 62062055. Youth Talents of Science and Technology in Universities of Inner Mongolia Autonomous Region under Grants NJYT24061. The Natural Science of Foundation of Inner Mongolia Autonomous Region under Grants JY20240061.}


\author[1,2]{Bo Li}[orcid=0009-0003-7011-4134]
\ead{20221800745@imut.edu.cn}

\credit{Conceptualization of this study, Methodology, Software, Validation, Data curation, Writing - original draft}

\affiliation[1]{organization={College of Data Science and Application},
                addressline={Inner Mongolia University of Technology}, 
                city={Huhhot},
                postcode={010080}, 
                state={Inner Mongolia},
                country={China}}

\affiliation[2]{organization={Inner Mongolia Autonomous Region Engineering and Technology Research Center of Big Data-Based Software Service},
                city={Huhhot},
                postcode={010080}, 
                state={Inner Mongolia},
                country={China}}

\affiliation[3]{organization={Inner Mongolia Key Laboratory of Beijiang Cyberspace Security},
                city={Huhhot},
                postcode={010080}, 
                state={Inner Mongolia},
                country={China}}

\author[1,2,3]{Jing Yun}[style=chinese]
\cormark[1]

\ead{yunjing_zoe@163.com}

\credit{Supervision, Writing - review \& editing}




\cortext[cor1]{Corresponding author  (Jing Yun).}


\begin{abstract}
Multi-view clustering can explore common semantics from multiple views and has received increasing attention in recent years. However, current methods focus on learning consistency in representation, neglecting the contribution of each view's complementarity aspect in representation learning. This limit poses a significant challenge in multi-view representation learning. This paper proposes a novel multi-view clustering framework that introduces a disentangled variational autoencoder that separates multi-view into shared and private information, i.e., consistency and complementarity information. We first learn informative and consistent representations by maximizing mutual information across different views through contrastive learning. This process will ignore complementary information. Then, we employ consistency inference constraints to explicitly utilize complementary information when attempting to seek the consistency of shared information across all views. Specifically, we perform a within-reconstruction using the private and shared information of each view and a cross-reconstruction using the shared information of all views. The dual consistency constraints are not only effective in improving the representation quality of data but also easy to extend to other scenarios, especially in complex multi-view scenes. This could be the first attempt to employ dual consistent constraint in a unified MVC theoretical framework. During the training procedure, the consistency and complementarity features are jointly optimized. Extensive experiments show that our method outperforms baseline methods.
\end{abstract}



    


\maketitle

\section{INTRODUCTION}
\label{sec1}
\begin{figure}[htbp]
\centering
\includegraphics[width=0.8 \linewidth]{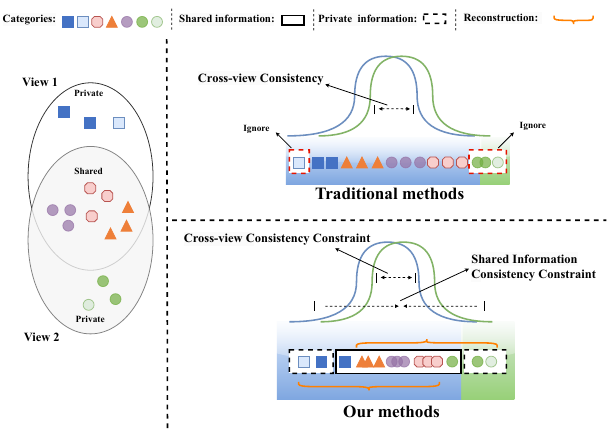}
\caption{Our basic idea. Taking a bi-view data as showcase, in left of figure, we use 2 oval panels to denote 2 views, polygons with different colors and shapes to indicate different categories. In right of figure, the solid and dotted rectangles denote the shared information and the private information. The red dotted rectangles denote ignoring the role of private information in multi-view clustering. The orange brackets indicate the reconstruction of multi-view data. Traditional methods improve clustering performance by pursuing multi-view consistency, which ignoring private information. However, when dual consistency constraints guarantee consistency learning, our method utilizes private information for view reconstruction. The goal of the cross-view consistency constraints is to maximize the mutual information between View 1 and View 2. The goal of the shared information consistency constraint is to express properly View 1 and View 2. Since complementarity information is specific to each view, multi-view representation learning requires explicitly preserving it to guarantee information completion during reconstruction.}
\label{fig_motivation}
\end{figure}
Multi-view data consists of data collected from different sources and is widely existent in real-world application scenarios \cite{cheng2018tensor, huang2020auto, nie2017multi, wu2019essential}. For example, we typically describe news events in terms of visual features and textual features that originate from different feature extractors.
To handle such heterogeneous features, multi-view clustering (MVC), as an important unsupervised paradigm for multi-view learning, aims to improve the learning of common semantics by mining the latent structures that are hidden in multi-view data to obtain better clustering performance. MVC is a fundamental task in the fields of data mining, pattern recognition, etc. 

It has received increasing attention in recent years \cite{fu2022low, li2021consensus, peng2018structured}. Consistency and complementarity are two key ingredients for boosting MVC. Consistency is able to analyze data from many perspectives and come up with the same result. In fact, a multi-view can accurately convey every aspect of an objective, but an individual view cannot. Thus, a comprehensive analysis leads to a more complete representation of the object. For example, in medical diagnosis and criminal analysis, complementary information affects the accuracy of judgment and is critical to decision-making. The critical factor to the success of the existing work is to learn consistency across different views. However, in the pursuit of consistency, complementary information is unavoidably ignored. Thus, multi-view clustering remains a challenging problem.

Traditional multi-view clustering methods are based on machine learning methods for clustering and can be divided roughly into three categories. Namely, (i) subspace-based clustering methods work on dimensionalizing high-dimensional, multi-view data and exploring its specific subspaces to learn shared and integrated subspace representations \cite{fu2022low, zhang2017latent}. Zhang et al. \cite{zhang2020adaptive} learn consistent information about samples by fusing representations of multi-view. Yang et al. \cite{yang2022hierarchical} proposed a subspace-based MVC hierarchical representation model with integration of intra-sample adaptive latent representation learning, intra-view global similarity structure learning, and inter-view structure-preserved consistency learning to address the problem of imprecise low-dimensional subspace representation and the inadequacy of exploring consistency. (ii) matrix factorization-based clustering methods factorize multi-view data into a consensus representation that is universally shared by all views \cite{wen2018incomplete, zhao2017multi, huang2020auto, liu2022efficient}. Cui et al. \cite{cui2019self} automatically weighted learned matrices for each view for clustering by deep factorization of multi-view matrices. (iii) graph-based clustering methods learn a consensus graph by mutual augmentation of nodes and edges \cite{li2021consensus, liu2022efficient, liu2022fast, wang2022align}. Nie et al. \cite{nie2017self} learn a multi-view clustering structure by exploring Laplace rank-constrained graphs. However, these methods suffer from poor representation capabilities and high computational complexity. The performance is limited in complex real-world data scenarios.

Due to the powerful fitting capabilities of deep neural networks, deep multi-view clustering methods have recently been extensively developed with increasingly superior feature representation capabilities. Li et al. \cite{li2019deep} incorporate features of all views to achieve a common representation of all views. Lin et al. \cite{lin2021completer} explored the common semantics of all views by dimensionalizing the features and employing a consistent learning objective. However, meaningless complementary information may dominate the feature learning process, thus interfering with the clustering quality. To reduce the influence of view complementarity information on clustering, several methods were designed for different alignment models \cite{du2021deep, trosten2021reconsidering, xu2022multi, zhou2020end}. For instance, aligning representation distributions and labeling distributions from different views by KL divergence \cite{hershey2007approximating}. They might be difficult to distinguish between clusters since the categories of one view might be aligned with different categories in another view. In addition, other methods aligned the representations from different views through contrastive learning \cite{lin2021completer, xu2022multi, lin2022dual, yang2022robust, wang2023self}. However, they usually need to achieve the reconstruction objective on the same features to avoid the trivial solution. Although the above methods achieve significant MVC performance improvement by considering the common semantics of learning consistency and reducing the effect of complementarity through alignment, the interference caused by entanglement of information among multi-view is also ignored. To be specific, consistency and complementarity information are entangled in a single feature space, which reduces the ability of the model to learn and reconstruct the representation data.

Based on the above observations, we found that cluster information is discrete, which is an abstraction of the common information among all views. And complementarity information is continuous, which has different effects on clustering. For instance, the observations from different perspectives are conducive to better describing the objects. Nevertheless, information is not always complementary to clustering and can even cause interference. In Fig. \ref{fig_motivation}, this information has a non-negligible role in reconstructing the object. Our objective is to disentangle consistent and complementary information and learn interpretable multi-view visual representations. Meantime, we hope that with consistency learning objectives assured, complementarity information is used to reconstruct the objective.

\begin{figure*}[htbp]
  \begin{center}
  \includegraphics[width=1\textwidth]{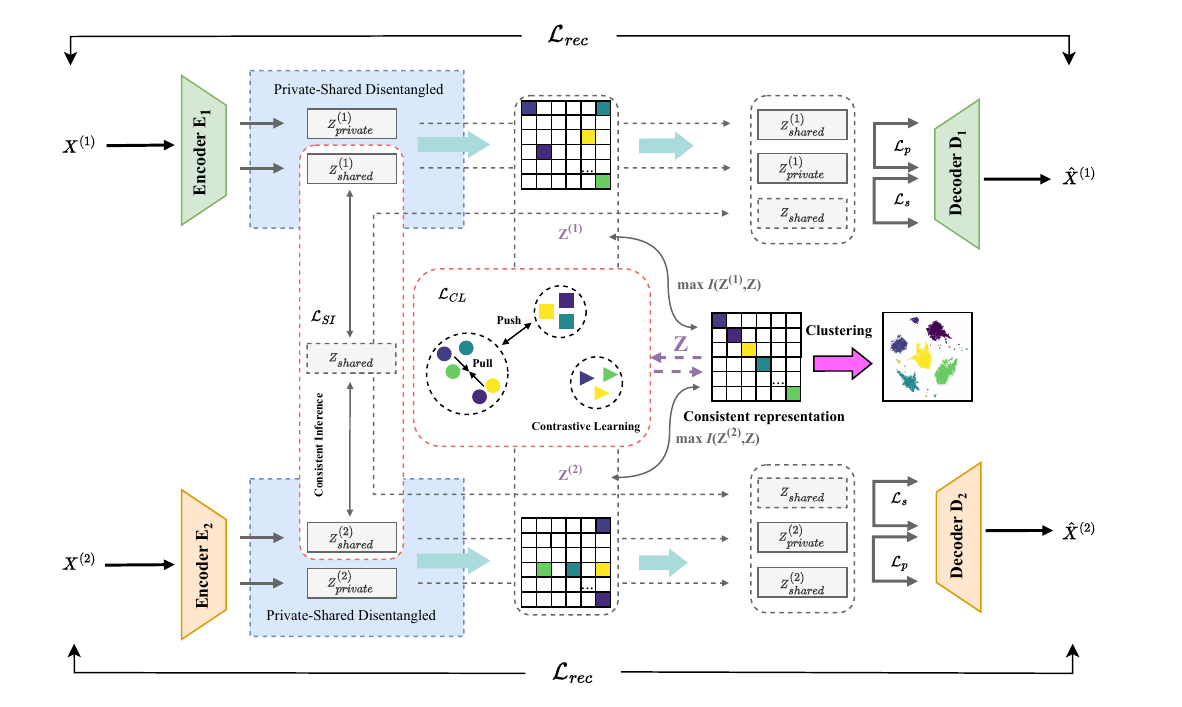}
  \caption{Overview of the framework. Bi-view data is used as a showcase in this figure. Our method consists of three joint learning objectives, i.e., reconstruction, variational reconstruction, shared information consistency inference constraint, and contrastive learning consistency constraint. Specifically, the goal of reconstruction is to maintain the diversity of views and project all views into latent spaces. Variational reconstruction consists of within-view reconstruction and cross-view reconstruction, which effectively ensure the quality of data representation. The shared information consistency inference objective ensures that the consistency of shared information across views in the latent space and aids in capturing the common features of the sample. Contrastive learning is to maximize the mutual information pursuit consistency among different views. This dual consistent constraint is interacted with and jointly optimized by the VAE network, thus improving the multi-view clustering performance.
  }
  \label{fig_framework}
  \end{center}
\end{figure*}

In this paper, as shown in Fig. \ref{fig_framework}, we propose a novel multi-view clustering framework (DCCMVC), which can learn disentangled and explainable representations, i.e., shared information and private information. On the one hand, due to the fact that the clustering information is discrete and the complementarity information is continuous, we choose the Gumbel Softmax distribution for the prior distribution of the consistency information and the Gaussian distribution for the complementarity information. By controlling the KL divergence between the latent representation and their prior during training, consistency and complementary information can be disentangled, which are further used for clustering tasks. On the other hand, since the representations of consistency information in multi-view are usually highly similar, we can maximize the mutual information and learn a consistent representation from different views by contrastive learning, which is useful for clustering tasks. Moreover, in an end-to-end learning framework combined with a VAE for consistency-complementarity disentanglement, the shared information of all views can be used to model the cross-view reconstruction between private information of each view, which is useful for reconstruction tasks. Therefore, the conflict between the reconstruction objective and consistency objectives is alleviated. Extensive experiments on eight widely used datasets demonstrate the effectiveness and superiority of our proposed DCCMVC compared to existing multi-view clustering methods. The main contributions of this paper are summarized as follows:
\begin{itemize}
    \item We propose a novel multi-view clustering framework of dual consistent constraint via disentangled consistency and complementarity for multi-view clustering (DCCMVC), which can separate the information of consistency and complementarity among multi-view data. Among them, consistency information is used for shared information consistency inference and cross-reconstruction, and complementarity information is used for within-view reconstruction.

    \item From the perspective of latent spaces, the proposed DCCMVC implements a dual consistency constraint. On the one hand, cross-reconstruction across several viewpoints is supported by the shared information consistency inference condition. Contrastive learning, on the other hand, is beneficial for learning consistency across views.
    
    \item We validate the effectiveness of DCCMVC by conducting extensive experiments on eight datasets. The experimental results show that our method provides superior performance over several state-of-the-art multi-view clustering methods.
\end{itemize}

\section{RELATED WORKS}
\label{sec2}

\subsection{Multi‐view Clustering.}
Recently, multi-view clustering has received a lot of attention \cite{chen2020multi,geng2022view,yang2019split,yin2021cauchy,zhang2019ae2,zhou2019dual}. In general, multi-view clustering methods can be broadly categorized into 5 classes due to different self-supervised signals. The first class is based on multi-view subspace clustering methods, which are based on the assumption that the features of multi-view share a common subspace and learn a consistent representation of the samples by fusing the features of multi-view. Cao et al. \cite{cao2015diversity} extended the traditional subspace clustering and proposed a diversity-inducing mechanism for multi-view subspace clustering. Yang et al. \cite{yang2022robust} recovered corrupted tensor data through latent low-rank tensor structures and accurately sought out the cluster structure through approximate representations and exploited the intrinsic subspace. Sun et al. \cite{sun2019self} seamlessly integrate spectral clustering and affinity learning in the context of deep learning. It utilizes clustering results as a guide to learn implicit representations for a single view and to build shared implicit subspaces across multiple views. The second class is based on matrix factorization, which explores the common latent factors of multi-view through matrix factorization. Cai et al. \cite{cai2013multi} introduced a shared metrics matrix for multi-view and handled the constraint matrix factorization problem. Zhao et al. \cite{zhao2017multi} learned a common feature representation through a semi-negative matrix decomposition, ensuring that the consensus representation retains most of the shared structural information between multiple graphs and has higher consistency in the hierarchy. In addation, Wei et al. \cite{wei2020multi} discretized the multi-view data matrix layer by layer through an asymptotic approach and generated a cluster at each layer to guide the clustering process. The third class is the graph-based methods, which address the issue by constraining the similarity matrices of each view and consensus graph matrices\cite{peng2019comic}. Fan et al. \cite{fan2020one2multi} introduce graph autoencoders that learn node embeddings based on a single information graph to capture view consistency from low-dimensional feature representations. Subsequently, Cheng et al. \cite{cheng2021multi} explored the consistency of high-dimensional samples in graph potential representations using a two-channel encoder. Recently, Xia et al. \cite{xia2022multi} imposed diagonal constraints on the consensus representation of multiple GCN autoencoders to improve clustering performance.
The fourth class is the pseudo label-based methods, which has shown to greatly improve clustering performance by designing agent tasks to implement pseudo-label learning, thus facilitating representation learning \cite{dubourvieux2021unsupervised, ge2020self, niu2022spice}. 
Given the high confidence inherent in pseudo-labeling, generating accurate pseudo-labels is crucial. Cui et al. \cite{cui2021self} used spectral clustering to generate pseudo-labels to guide multi-view fusion, which facilitates clustering. In addition, they propose an alternative iterative optimization algorithm to rationalize pseudo-labels efficiently.
The fifth class is the complementarity-based methods. Since different views often contain complementary information. To obtain a more comprehensive and accurate description of data objects, complementary information can be utilized to enhance internal clustering across views. 
Existing MVC approaches typically investigate the complementarity of multi-view data through a fusion process.
Wang et al. \cite{wang2021generative} utilized complementary information between different views through adaptive weighted fusion and used a learned generic representation to help interpolate the data. Although fusion is an effective method, it can negatively affect clustering due to low data quality and redundancy of complementary information. Li et al. \cite{li2024balancing} balance the consistency and complementarity information in multi-view by delayed activation to enhance the clustering performance.

In addition, generative class-based MVC methods map data to potential space for representation learning or incomplete data recovery through generative adversarial networks, variational autoencoders, diffusion models, etc., to boost clustering. Wang et al. \cite{wang2023self} recover incomplete data by generating adversarial networks to learn more complementary and consistent potential representations. Xu et al. \cite{xu2021multi} efficiently fused consistency and complementarity features via a variational autoencoder.
Meanwhile, contrastive class-based MVC methods are used to facilitate clustering by contrastive learning in order to efficiently distinguish features. Yang et al. \cite{yang2023dealmvc} constrained feature consistency by double contrastive loss locally and globally. Lin et al. \cite{lin2021completer} learned information-rich and consistent representations by maximizing the mutual information between different views through contrastive learning. Yang et al. \cite{yang2022robust} propose a noise-robust contrastive loss that mitigates or even eliminates the effects of false negatives during random sampling of negative samples. Benefiting from the powerful feature representation of deep networks, deep multi-view clustering has the ability to extract finer feature representations that can more effectively reveal potential clustering patterns in multi-view data.

\subsection{Variational Autoencoder.}
Recently, autoencoders (AE) have shown excellent performance in high-dimensional data representation \cite{guo2018deep,ren2019semi}. The combination of variational inference and autoencoders leads to the creation of variational autoencoders (VAE) \cite{kingma2013auto}. A series of VAE-based multi-view or multi-modal learning methods have emerged \cite{federici2020learning,lee2020private,wu2018multimodal,yin2020shared}. Jiang et al. \cite{jiang2017variational} first proposed a VAE-based deep clustering framework, which uses Gaussian mixture models for data generation. Yang et al. \cite{yang2019deep} proposed graph embedding in Gaussian mixture of variational autoencoders. Compared to ordinary representation learning, the goal of disentanglement is to obtain explainable factors hidden in the data \cite{bengio2013representation}. InfoGAN \cite{chen2016infogan} and $\beta$-VAE \cite{higgins2017beta} are two famous unsupervised untangling methods. However, InfoGAN generates samples with limited quality. To alleviate the problem, Dupont et al. \cite{dupont2018learning} separated discrete clustered information from continuous complementary information based on the VAE framework. Burgess et al. \cite{burgess2018understanding} further improved the reconstruction quality and untangling by increasing the upper limit of KL divergence.

Existing work considers the use of VAE for modeling multi-modal data. Suzuki et al. \cite{suzuki2016joint} used VAE to implement the interaction of two modalities to solve the modal absence problem. Shi et al. \cite{shi2019variational} applied Mixture-of-Expert (MoE) to jointly learn shared factors across multiple modalities. This paper attempts to separate features into shared and private information with a disentanglement VAE, which can improve the quality of reconstruction and the effectiveness of consistency in representation learning.

\subsection{Contrastive Learning.}
Contrastive learning is an unsupervised representation learning method. Recently, it has achieved promising performance in computer vision. Its goal is to maximize the similarity of positive pairs while minimizing the similarity of negative pairs in the feature space \cite{yang2022robust,zeng2023semantic,li2021contrastive,lin2022dual}. The initial method of minimizing InfoNCE loss was introduced \cite{van2018representation} by explicitly defining pairs of positive and negative samples to assess dissimilarity between samples. CMC \cite{hassani2020contrastive}, MoCo \cite{he2020momentum} and SimCLR \cite{chen2020simple} all these by minimizing the InfoNCE loss function. Specifically, CMC treats several different views of an image as positive samples while designating another view as a negative sample. MoCo and SimCLR emphasize the importance of a positive sampling strategy. In contrast, to avoid the definition of negative sample pairs, BYOL \cite{grill2020bootstrap}, SimSiam \cite{chen2021exploring}, and DINO \cite{caron2021emerging} abandon negative sample pairs altogether and successfully convert the contrastion task into a prediction task. Similarly, SwAV \cite{caron2020unsupervised} promotes consistency of the same sample in different views. Ultimately, it achieved excellent results. 

Existing contrastive learning methods push the distance between anchor points and negative samples farther apart while decreasing the distance between anchor points and positive samples, regardless of the distance between differently labeled data in the negative sample set. To deal with the fact that the distance between negative samples with different labels may inadvertently decrease while the distance between the anchor point and the negative sample increases, Li et al. \cite{li2025triple} separated negative samples from the same anchor point by triple contrast learning. Although existing studies have shown that consistency could be learned by maximizing mutual information, they ignore that catastrophic forgetting of features at different stages. Recently, Zhang et al. \cite{zhang2025learning} learned unlearned features incrementally by capturing biased feature distributions at each stage with negative sampling and then preserved the features by integrating them across stage representations. In brief, contrastive learning utilizes the variability between samples to extract rich self-supervised signals to help model training, especially in large-scale data scenarios, which can greatly improve the quality of representation learning and has already demonstrated its research value in several research areas \cite{zhang2025disentangled, mo2025multi, cui2025migcl, huang2025extracting}.

\section{METHOD}
\label{sec3}
In this paper, a multi-view data, which includes $N$ samples with $V$ views, is denoted as $\left \{ X_{1}^{v}; \cdots; X_{N}^{v} \right \} \in \mathcal{R}^{N \times D_{v}}$, where $D_v$ is the dimension of the feature in the $v$-th view. Multi-view clustering aims to divide the samples into $K$ clusters. This section proposes a disentangled consistency and complementarity and dual consistent constraints network for multi-view clustering, termed DCCMVC. The framework is shown in Fig. \ref{fig_framework}. Our motivation is built on variational autoencoder (VAE), and we first emphasize the relevant aspects of the VAE-based model and then use it to structure our method. In contrast to ordinary representation learning, disentangled representation learning aims to obtain explainable factors hidden in data \cite{bengio2013representation}. Our method consists of three main components: consistency and complementarity disentangled, consistency information inference, and cross-view consistency. For clarity, we will first describe each part in detail and finally summarize the final objective function.

\textbf{Variational Autoencoder in Multi-view.}

\begin{figure}[t!]
\centering
\includegraphics[width=1\linewidth]{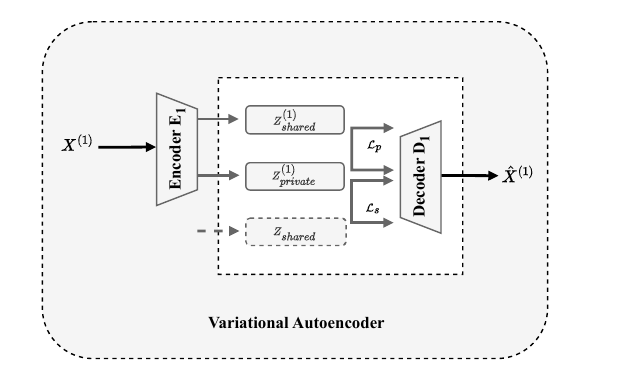}
\caption{The structure of variational autoencoder (VAE).
}
\label{fig_vae}
\end{figure}

With the concept of multi-view description, the latent space can be divided into a private space (complementarity) for each view and a shared space (consistency) across all views. Our goal is to obtain separation-friendly private and shared spaces. Shared spaces can only transfer shared information across views and cannot reconstruct aspects of a view. However, private information can achieve high fidelity in reconstruction on the one hand and promote the separation of private information across views on the other. Therefore, as shown in Fig. \ref{fig_vae}, it is essential to the reconstruction of private information in generative models such as VAE.
We introduce a VAE-based generative model:
\begin{equation}
\label{eq1}
\begin{split}
p(X,Z_p,Z_s)=p(X|Z_p,Z_s)p(Z_p,Z_s),
\end{split}
\end{equation}
where $Z_s$ denotes the shared information across all views and $Z_p$ is the private information for each view. 

We assume that the latent representation can be decomposed into $\left \{ Z_{p}, Z_{s} \right \}$. 
In general, given multi-view dataset $\left\{ X^{(1)}, X^{(2)}, \cdots,X^{(N)} \right \}$, the latent representation of each view can be described as $E_{\Phi _i}(Z_{p},Z_{s} \mid X^{(i)})$. Here $\Phi _i$ is the parameter learned by the variational encoder $E$ for each view. We describe how to achieve disentanglement of shared and private information in Sec. \ref{sec3.1}. We seek efficient constraint to shared information consistency inference across all views, and we describe how to make the shared information  $p(Z_{s} \mid X^{(1)},X^{(2)})$ approximate its prior in Sec. \ref{sec3.2}. In addition, we introduce contrastive learning pursuing consistency across views in Sec. \ref{sec3.3}

\subsection{Consistency and Complementarity  Information Disentangled in Multi-view}
\label{sec3.1}
The variational autoencoder (VAE) implements variational inference on latent variables by autoencoder structures. The objective is to maximize the marginal distribution $p(X)= \int p_{\theta}(X \mid Z )p(Z)dX$, i.e., the likelihood function for multi-view view data, but it is difficult to compute the true joint posterior $p_{\theta}(X \mid Z )$. Therefore, we maximize $\mathbb E_{z \sim q_{\Phi}(z \mid x)} \left[ log p_{\theta (x \mid z)} \right]$ using the approximate identification model $q_{\Phi}(z \mid x)$ by introducing the evidence lower bound (ELBO) for each view while minimizing $KL(q_{\Phi}(z\mid x),p(z))$ to replace the true posterior. The ELBO for each view can be defined as:
\begin{equation}
\label{eq2}
\begin{split}
\mathcal L_{ELBO}(X)
&=\mathbb E_{q_{\Phi}(Z_{p},Z_{s}\mid X)}[log p_{\theta}(X \mid Z_{p},Z_{s})]\\
&-D_{KL}(q_{\Phi}(Z_{p}, Z_{s}\mid X) \parallel p(Z_{p}, Z_{s}))  \le logp(X),
\end{split}
\end{equation}
where $q_{\Phi}(Z_{p}, Z_{s}\mid X)$ and $p_{\theta}(X\mid Z_{p},Z_{s})$ and denote the encoder and decoder used to learn the parameters $\Phi$ and $\theta$, respectively. The first term $\mathbb E_{q_{\Phi}(Z_{p},Z_{s}\mid X)}[log p_{\theta}(X \mid Z_{p},Z_{s})]$ of ELBO (Eq. (\ref{eq2})) is optimized to learn both shared and private information to achieve disentanglement, namely VAE reconstruction loss. The second term $D_{KL}(q_{\Phi}(Z_{p}, Z_{s}\mid X) \parallel p(Z_{p}, Z_{s}))$ plays the role of regularization, which aims to approach the prior distribution $p(Z_{p},Z_{s})$, and is optimized to ensure the reconstruction quality of the VAE.

Next, we discuss in more detail the role of the second term on the disentanglement. The assumption of this paper is that the shared and private information of the views are conditionally independent, i.e., $q_{\Phi}(Z_{p}, Z_{s}\mid X)=q_{\Phi}(Z_p \mid X)q(Z_s \mid X)$ and $p_{}(Z_{p}, Z_{s})=p(Z_p)p(Z_s)$ and the KL divergence ($D_{KL}$) can be decomposed as:
\begin{equation}
\label{eq3}
\begin{split}
&D_{KL}(q_{\Phi}(Z_{p}, Z_{s}\mid X) \parallel p(Z_{p}, Z_{s}))\\
&=\mathbb E_{q(Z_p,Z_s \mid X)}\left[log \frac{q(Z_p,Z_s\mid X)}{p(Z_p,Z_s)}\right] \\
&=\mathbb E_{q(Z_p \mid X)}E_{q(Z_s \mid X)}\left[log \frac{q(Z_p\mid X)q(Z_s \mid X)}{p(Z_p)p(Z_s)}\right] \\
&=\mathbb E_{q(Z_p \mid X)}E_{q(Z_s \mid X)}\left[log \frac{q(Z_p\mid X)}{p(Z_p)}\right] \\
&+ \mathbb E_{q(Z_p \mid X)}E_{q(Z_s \mid X)}\left[log \frac{q(Z_s\mid X)}{p(Z_s)}\right]\\
&=D_{KL}(q(Z_p\mid X)\parallel p(Z_p)) + D_{KL}(q(Z_s\mid X)\parallel p(Z_s)),
\end{split}
\end{equation}
In this way, the KL divergence terms for shared and private information are separated, which is designed for disentangling the view-shared and view private representations. For the $v$-th view, the objective to be maximized becomes:
\begin{equation}
\label{eq4}
\begin{split}
&\mathcal L_{ELBO}(X) =\mathbb E_{q_{\Phi}(Z_{p},Z_{s}\mid X)}[log p_{\theta}(X \mid Z_{p},Z_{s})]\\
&- D_{KL}(q(Z_p\mid X)\parallel p(Z_p)) - D_{KL}(q(Z_s\mid X)\parallel p(Z_s)),
\end{split}
\end{equation}
where the first term is the shared and private information learned by disentanglement.
The second term is the KL divergence of the private information, denoting the difference between the posterior distribution $q_{}(Z_{p}^{}\mid X^{})$ and the prior distribution $p(Z_{p}^{})$.
The third term is the KL divergence of the shared information, which represents the difference between the posterior distribution $q_{}(Z_{s}^{}\mid X^{})$ and the prior distribution $p(Z_{s})$.

\begin{figure}[t!]
\centering
\includegraphics[width=1\linewidth]{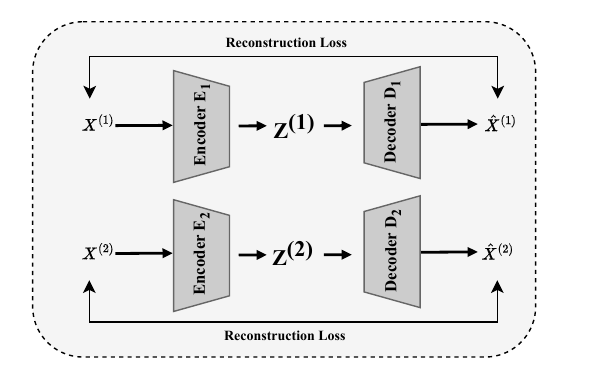}
\caption{The process of view reconstruction.
}
\label{fig_rec}
\end{figure}

Considering the effect of two factors, one is that different views have different feature dimensions in which feature redundancy and random noise are usually present, and the other is that different views have different data reconstruction loss. To mitigate the effect of negative factors on clustering, we employ an autoencoder to compress the original features to the same dimensions to obtain a feature representation of the samples, as shown in Fig \ref{fig_rec}. Meanwhile, the corresponding decoder is used to reconstruct the original view to ensure that the network learns the original feature information completely. Thus, we pass each view to the autoencoder. The introduced view reconstruction loss $\mathcal L_{rec}$ is represented as follows:
\begin{equation}
\label{eq5}
\begin{split}
\mathcal L_{rec}
&=\sum_{v=1}^{N} \left \| X^{(v)} -\hat{X}^{(v)} \right \| _2^2\\
&=\sum_{v=1}^{N} \left \| X^{(v)} -D_v(E_v(Z_p^{(v)},Z_s \mid X^{(v)})) \right \| _2^2,
\end{split}
\end{equation}
where $E_v$ and $D_v$ denote the encoder and decoder of the $v$-th view. We alleviate the missing view feature information by introducing a reconstruction loss and avoiding trivial solutions.

\subsection{Consistency Information Inference in Latent Space}
\label{sec3.2}
Given multi-view data $X = \left \{X^{(1)},X^{(2)}, \dots,X^{(N)} \right \}$, we first define the latent space inference, i.e., each view has a posterior distribution $p(Z^i \mid X^{(i)})$, which is approximated by $q(Z_{p}^{i},Z_{s}^{i}\mid X^{(i)})$. Therefore, we require that the shared information has a similar distribution across all views to ensure that the shared information has a consistent representation, i.e., $Z_{s}^{i}=Z_{s}, w.p.1,\forall i$. We separate the private information $q(Z_{p}^{i} \mid X^{(i)})$ from the shared information $q(Z_{s} \mid X)$ and the approximate a posteriori shared information $q(Z_{s}\mid X)$ is defined as follows: 
\begin{equation}
\label{eq6}
\begin{split}
q(Z_{s} \mid X) =  \prod_{i=1}^{N} q(Z_{s} \mid X^{(i)}),
\end{split}
\end{equation}

\begin{figure}[t!]
\centering
\includegraphics[width=0.6\linewidth]{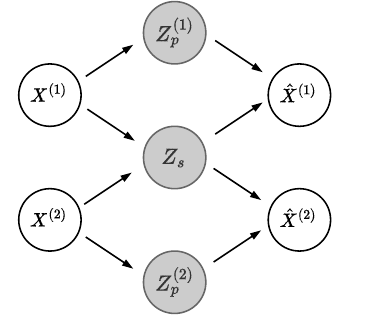}
\caption{The process of shared information consistency inference.
$Z_{p}^{(1)}$ and $Z_{p}^{(2)}$ denotes the private information of view $X^{(1)}$ and view $X^{(2)}$, respectively. $Z_{s}$ denotes the shared information through consistency inference constraint. The $\hat X^{(1)}$ and $\hat X^{(2)}$ denotes the reconstructed view by the private information of each view and the shared information of all views. We elaborate on the shared information consistency inference constraint in Eq. \ref{eq10}.
}
\label{fig_inference}
\end{figure}

According to the Gumbel-Max reparameterization trick \cite{gumbel1954statistical}, we can further get the following expression:
\begin{equation}
\label{eq7}
\begin{split}
q(Z_s \mid X) = \mathcal G(S) = \frac{exp((log S_i +g_i)/ \tau )}{\sum_{m=1}^{N}exp((log S_m + g_m)/ \tau)},
\end{split}
\end{equation}
where $g \sim \mathcal G(0,1)$ and $\tau$ is the temperature parameters that control the relaxation. In addition, the prior for the private information of the view is the standard normal distribution, i.e., $Z_p \sim \mathcal N(0, I)$. $q(Z_p \mid X)$ is parameterized by the factorization Gaussian:
\begin{equation}
\label{eq8}
\begin{split}
q(Z_p \mid X) = \prod_{i=1}^{N} q(Z_{p} \mid X^{(i)}),
\end{split}
\end{equation}
According to the reparameterization trick \cite{kingma2013auto, rezende2014stochastic}, we have the following terms can be equated:
\begin{equation}
\label{eq9}
\begin{split}
q(Z_p \mid X) = \mathcal N(\mu_, \delta ^{2}),
\end{split}
\end{equation}
As shown in Fig. \ref{fig_inference}, each view is disentangled into shared information $Z_s$ and private information $Z_p$.  The shared information through consistency inference constraint can effectively help cross-view reconstruction. Specifically, given a view $X^{(1)}$, the view $X^{(2)}$ is realized by using the prior $Z_{p}^{(2)}$ and the prior distribution of $Z_{s}$ with $X^{(2)}$ private information, and vice versa for the view $X^{(1)}$. The loss of the shared information consistency inference process $\mathcal{L}_{SI}$ can be expressed as follows:
\begin{equation}
\label{eq10}
\begin{split}
\mathcal{L}_{SI}
&=p(\hat X^{(2)}\mid X^{(1)}) \\
&= \mathbb E_{p(Z_{p}^{(2)})q(Z_{s} \mid X^{(1)})}\left[p(\hat X^{(2)}\mid Z_{p}^{(2)},Z_{s}) \right],
\end{split}
\end{equation}

Based on the latent representation of disentangled and inferenced, a critical issue is: how do we achieve the shared information and private for reconstruction? We attempt to learn variational joint posterior distributions of shared information to synthesize from one view to another under different reconstruction paradigms. It is noted that the private information of each view may be complementary for clustering, but it's certainly useful for reconstruction.

Given the private information $Z^{i}_{p}$ and shared information $Z^{i}_{s}$ of each view $X^{(i)}$, we first conduct within-view reconstruction to generate the expected representation of each view as follows:
\begin{equation}
\label{eq11}
\begin{split}
\mathcal L_{private} = \mathbb E_{q(Z_{p}^{i} \mid X^{(i)})q(Z_{s}^{i} \mid X^{(i)})}\left[p(\hat X^{(i)} \mid Z_{p}^{i},Z_{s}^{i}) \right],
\end{split}
\end{equation}
Then, we model the accuracy for within-view reconstruction to compensate with the KL divergence from the prior. The within-view reconstruction loss $\mathcal L_{p}$ can express as:
\begin{equation}
\label{eq12}
\begin{split}
&\mathcal L_{p} = \sum_i \mathbb E_{p(X^{(i)})} \{ \varepsilon \mathcal L_{private} \\
&-D_{KL}(q_{\Phi}(Z_{p}^{i}\mid X^{(i)})\parallel (p(Z_{p}^{i})) \\
&- D_{KL}(q_{\Phi}(Z_{s}^{i} \mid X^{(i)})\parallel p(Z_{s}^{i})) \},
\end{split}
\end{equation}
where $\varepsilon$ balances the reconstruct across different views. The first term $\mathcal L_{private} $ is the within-view reconstruction loss, which denotes reconstructing the expectation of the log-likelihood of each view $X^{(i)}$ under the premise that the private information $Z^{i}_{p}$ and the shared information $Z_{s}$ are obtained after disentanglement of each view. The second term $D_{KL}(q_{\Phi}(Z_{p}^{i}\mid X^{(i)})\parallel (p(Z_{p}^{i}))$ is the KL divergence of private information of each view, which denotes the difference between the posterior distribution $q_{\Phi}(Z_{p}^{i}\mid X^{(i)})$ and the prior distribution $p(Z_{p}^{i})$. The third term $D_{KL}(q_{\Phi}(Z_{s}^{i} \mid X^{(i)})\parallel p(Z_{s}^{i}))$ is the KL divergence of shared information of each view, which denotes the difference of each view between the posterior distribution $q_{\Phi}(Z_{s}^{i}\mid X^{(i)})$ and the prior distribution $p(Z_{s}^{i})$.

Next, we conduct cross-view reconstruction using private information of each view and shared information of all views, generating an expected representation for each view. The cross-view reconstruction process can be express as:
\begin{equation}
\label{eq13}
\begin{split}
\mathcal{L}_{shared} = \mathbb E_{q(Z_{p}^{i} \mid X^{(i)})q(Z_{s}\mid X)}\left[p(\hat X^{(i)} \mid Z_{p}^{i},Z_{s}) \right],
\end{split}
\end{equation}
Finally, we model the accuracy for cross-view reconstruction is again compensate by KL divergence from the prior. The cross-view reconstruction loss $\mathcal L_{s}$ can be express as:
\begin{equation}
\label{eq14}
\begin{split}
&\mathcal L_{s} = \sum_i \sum_j \{ \omega \mathcal{L}_{shared} \\
&-D_{KL}(q_{\Phi}(Z_{p}^{i} \mid X^{(i)})\parallel p(Z_{p}^{i}))\\
&-D_{KL}(q_{\Phi}(Z_{s} \mid X^{(j)})\parallel p(Z_{s}))\},
\end{split}
\end{equation}
where $\omega$ balances the reconstruction across different views again. The first term $\mathcal{L}_{shared}$ denotes the cross-view reconstruction loss. The second term $D_{KL}(q_{\Phi}(Z_{p}^{i} \mid X^{(i)})\parallel p(Z_{p}^{i}))$ denotes the KL divergence of private information of each view. The third term $D_{KL}(q_{\Phi}(Z_{s} \mid X^{(j)})\parallel p(Z_{s})))$ denotes the KL divergence of shared information of all views.

\subsection{Cross-view Consistency Learning}
\label{sec3.3}
\begin{figure}[t!]
\centering
\includegraphics[width=1\linewidth]{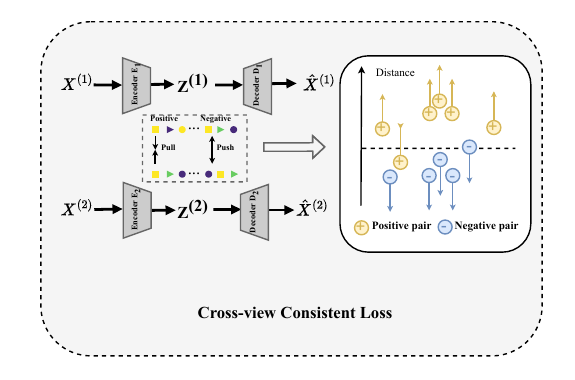}
\caption{The process of consistency learning.
}
\label{fig4}
\end{figure}

As indicated above, we constrain the shared information  consistency, which ensures the quality of both within-view reconstruction and cross-view reconstruction. To enhance the capability of learning consistency across different views, as shown in Fig. \ref{fig4}, we introduce contrastive learning in the parameterized latent space to learn the latent representation across views. Specifically, we concatenate the private and shared information of the view into a feature, i.e., $Z^{(i)}=\left \{Z_p^{(i)},Z_s^{(i)} \right\}$. Naturally, we assume that features from different views of the same sample should have closer proximity in the latent space and that features from different samples should be farther apart. Inspired by \cite{lin2021completer}, we bound by maximizing a lower bound on the mutual information, which further ensures that the representations learned through within-view reconstruction are discriminative representations of the samples. The corresponding loss $\mathcal L_{CL}$ representation is as follows:
\begin{equation}
\label{eq15}
\begin{split}
\mathcal L_{CL} = - \sum_{}^{}(I(Z^{(1)},Z^{(2)})+\alpha(H(Z^{(1)})+H(Z^{(2)}))),
\end{split}
\end{equation}
where $H$ denotes the information entropy, and a larger $H$ means $Z$ contains more information. In addition, maximizing $H(Z^{(1)}$ and $H(Z^{(2)})$ avoids generating simple solutions that assign all samples to the same cluster. $I$ denotes the mutual information, $Z^{(1)}$ and $Z^{(2)}$ denotes two discrete cluster assignment variables, and the inputs to the softmax function in the decoder are used as the cross-cluster probability distribution of $Z^{*}$ and to obtain the joint probability distribution of $Z^{(1)}$ and $Z^{(2)}$. The formula is as follows:
\begin{equation}
\label{eq16}
\begin{split}
I(Z^{(1)},Z^{(2)})= P_{Z^{(1)},Z^{(2)}} \frac{log(P_{Z^{(1)}}P_{Z^{(2)}})}{log(P^{\eta+1}_{Z^{(1)}}P^{\eta+1}_{Z^{(2)}})},
\end{split}
\end{equation}
where $\eta$ is the regularization factor.

\subsection{Learning Objective}
\label{sec3.4}
Overall, The optimization of DCCMVC is summarized in Algorithm. \ref{alg:cap}. The learning objectives of the proposed method can be formulated as follows:
\begin{equation}
\label{eq17}
\begin{split}
\mathcal L = \alpha \mathcal  L_{rec} + \beta (\mathcal{L}_{s} + \mathcal{L}_{p} + \mathcal{L}_{SI} )+ \gamma \mathcal L_{CL},
\end{split}
\end{equation}
where $\mathcal L_{rec}$ denotes the single view autoencoder reconstruction loss, $\mathcal L_{s}$ denotes cross-view reconstruction of the view using shared information of all views and private information of each view, $\mathcal L_{p}$ denotes within-view reconstruction of view using shared information and private information of each view, $\mathcal L_{SI}$ denotes the consistency constraint of shared information of all views, and $\mathcal L_{CI}$ denotes the consistency of constraint among all views. $\alpha, \beta$, and $\gamma$ are balancing factors for $\mathcal L_{rec}$, $\mathcal L_{s} + \mathcal L_{p} + \mathcal L_{SI}$, and $\mathcal L_{CI}$, respectively.

\begin{algorithm}
\caption{Optimization algorithm of DCCMVC}
\label{alg:cap}
\begin{algorithmic}
\Require 
\State Multi-view dataset $X^{(v)}=\left \{ X^{(1)},X^{(2)},\cdots, X^{(n)} \right \}$;
\State Learning rate $\eta$; Cluster number $K$;
\State Hyper-parameter $\alpha$, $\beta$, $\gamma$; 
\Ensure
\State Randomly initialize parameters.
\While{Pre-trained}
    \State Calculate reconstruction loss by Eq. (\ref{eq5}).
\EndWhile
\While{Trained}
    \State Disentangled $Z_s$ and $Z_p$ by encoders.
    \State Inference $q(Z_s \mid X)$ by Eq.(\ref{eq7}) and $q(Z_p \mid X)$ by Eq.(\ref{eq9}) with reparameterization tricks.
    \State The share information consistency inference constraint by Eq. (\ref{eq10}).
    \State Generate $p(X^{(i)} \mid Z_p^{i}, Z_s^{i})$ by Eq. (\ref{eq12}).
    \State Generate $p(X^{(i)} \mid Z_p^{i}, Z_s)$ by Eq. (\ref{eq14}).
    \State Cross-view consistency constraint by Eq. (\ref{eq15}).
\EndWhile
\State \textbf{Output:}
\State Obtain Consistent Feature Representation $Z_n$.
\State Obtain Clustering Results $R$.
\State \textbf{Return} $R$.
\end{algorithmic}
\end{algorithm}


\section{EXPERIMENTS}
\label{sec4}

\subsection{Experimental Settings}
\label{sec4.1}

\textbf{Datasets.}
The experiments are conducted on eight benchmark multi-view datasets. We summarise the details of the datasets in Table. \ref{table1}.

\begin{itemize}
    \item \textbf{BBCsport} \cite{greene2006practical}: contains 544 sports news articles from the BBC Sport website, which cover 2 views of 5 themes in 2004 and 2005, with 3,183 features in the first view and 3,203 features in the second view. 

    \item \textbf{Columbia Consumer Video (CCV)} \cite{jiang2011consumer}: is a video dataset containing 6,773 samples belonging to 20 categories with hand-crafted Bag-of-Words representations in STIP, SIFT and MFCC views.
    
    \item \textbf{MNIST-USPS} \cite{peng2019comic} : is a handwritten digit dataset, which contains 5,000 samples belonging to 10 categories with two different styles of digital images.

    \item \textbf{Reuters} \cite{yang2023dealmvc} : is a news dataset published by Reuters, which contains 18,758 samples belong to 10 topics.

    \item \textbf{Caltech} \cite{fei2004learning} collected 1400 images belonging to 7 categories with 5 views. Here four sub-datasets were constructed to evaluate the robustness of the comparison methods with respect to the number of views, namely, \textbf{Caltech-2V}, \textbf{Caltech-3V}, \textbf{Caltech-4V} and \textbf{Caltech-5V}. Specifically, Caltech-2V uses WM and centrist; Caltech-3V uses WM, CENTRIST and Lycium barbarum polysaccharide; Caltech-4V uses WM, CENTRIST, LGB and GIST; Caltech-5V uses WM, CENTRIST, LGB, GIST and HOG.
    
\end{itemize}

\begin{table}[htbp]
\renewcommand\arraystretch{1.3}
\centering
\footnotesize
\caption{The information of the datasets in our experiments.}
\label{table1}
\setlength{\tabcolsep}{0.1mm}{
\begin{tabular}{lcccc}
\hline
Datasets      & \# Category & \# View & \# Samples & \# Features \\ \hline
BBCSport & 5          & 2       & 544      & 3,183/3,203   \\
CCV      & 20          & 3       & 6,773      & 5,000/5,000/4,000       \\
Reuters      & 10         & 2       & 18,758      & 10/10      \\
MNIST-USPS      & 10          & 2       & 5,000      & 256/256       \\
Caltech-2V    & 7          & 2       & 1,400      & 40/254       \\
Caltech-3V    & 7          & 3       & 1,400      & 40/254/928       \\
Caltech-4V    & 7          & 4       & 1,400      & 40/254/928/512       \\
Caltech-5V    & 7          & 5       & 1,400     & 40/254/928/512/1,984     \\ \hline
\end{tabular}
}
\end{table}

\textbf{Compared methods.}
The proposed DCCMVC is the benchmarked against 6 prominent deep multi-view clustering methods, including COMIC \cite{peng2019comic}, SiMVC \cite{trosten2021reconsidering}, DSMVC \cite{tang2022deep}, CoMVC \cite{trosten2021reconsidering}, MFLVC \cite{xu2022multi}, DealMVC \cite{yang2023dealmvc}.

To further verify our methods, we followed the \cite{xu2022multi} setup and compared it with 12 other methods on Caltech-2V, 3V, 4V, and 5V. These methods are AE$^2$NET \cite{zhang2019ae2}, RMSL \cite{li2019reciprocal}, MVC-LFA \cite{wang2019multi}, CDIMC-net \cite{wen2021structural}, EAMC \cite{zhou2020end}, IMVTSC-MVI \cite{wen2021unified}, DEMVC \cite{xu2021deep}, CONAN \cite{ke2021conan}, CUMRL \cite{zheng2021collaborative}, SDMVC \cite{xu2022self}, CMIB \cite{wan2021multi}, and GUMRL \cite{zheng2022graph}, respectively.

\begin{table*}[ht!]
\caption{The clustering performance across eight multi-view benchmark datasets. The most exceptional results are marked in \textbf{bold}, and the second-best results are \underline{underlined}. The ‘-’ indicates unavailable results due to being out of memory.}
\label{table2}
\renewcommand\arraystretch{1.4}
\centering
\footnotesize
\setlength{\tabcolsep}{2.3mm}{

\begin{tabular}{lcccccccccccc}
\hline
Datasets & \multicolumn{3}{c}{BBCSport} & \multicolumn{3}{c}{CCV} & \multicolumn{3}{c}{MNIST-USPS} & \multicolumn{3}{c}{Reuters} \\ \cline{2-13} 
Metrics & ACC & NMI & PUR & ACC & NMI & PUR & ACC & NMI & PUR & ACC & NMI & PUR \\ \hline
(2019) COMIC \cite{peng2019comic} & 0.322 & 0.215 & 0.356 & 0.157 & 0.081 & 0.157 & 0.482 & 0.709 & 0.531 & 0.166 & 0.125 & 0.166 \\
(2021) CoMVC \cite{trosten2021reconsidering}& 0.375 & 0.198 & 0.408 & 0.296 & 0.286 & 0.297 & 0.987 & 0.976 & 0.989 & 0.322 & 0.135 & 0.338 \\
(2021) SiMVC \cite{trosten2021reconsidering} & 0.378 & 0.109 & 0.400 & 0.151 & 0.125 & 0.216 & 0.981 & 0.962 & 0.981 & 0.335 & 0.103 & 0.336 \\
(2022) DSMVC \cite{tang2022deep} & 0.419 & 0.146 & 0.475 & - & - & - & - & - & - & 0.438 & 0.181 & 0.450 \\
(2022) MFLVC \cite{xu2022multi} & 0.409 & 0.277 & 0.614 & \underline{0.312} & \underline{0.316} & \underline{0.339} & \underline{0.995} & \underline{0.986} & \underline{0.989} & 0.399 & 0.200 & 0.414 \\
(2023) DealMVC \cite{yang2023dealmvc}& \underline{0.807} & \underline{0.655} & \underline{0.807} & 0.290 & 0.302 & 0.331 & 0.795 & 0.902 & 0.790 & \textbf{0.470} & \underline{0.263} & \underline{0.483} \\
DCCMVC(Ours) & \textbf{0.889} & \textbf{0.732} & \textbf{0.843} & \textbf{0.359} & \textbf{0.323} & \textbf{0.371} & \textbf{0.995} & \textbf{0.993} & \textbf{0.995} & \underline{0.465} & \textbf{0.404} &\textbf{ 0.761} \\ \hline
\end{tabular}

}
\end{table*}

\begin{table*}[ht!]
\caption{The clustering performance 
across eight multi-view benchmark datasets. The most exceptional results are marked in \textbf{bold}, and the second-best results are \underline{underlined}. The ‘-’ indicates unavailable results due to being out of memory.}
\label{table3}
\renewcommand\arraystretch{1.4}
\centering
\footnotesize
\setlength{\tabcolsep}{2.1mm}{

\begin{tabular}{lcccccccccccc}
\hline
Datasets & \multicolumn{3}{c}{Caltech-2V} & \multicolumn{3}{c}{Caltech-3V} & \multicolumn{3}{c}{Caltech-4V} & \multicolumn{3}{c}{Caltech-5V} \\ \cline{2-13} 
Metrics & ACC & NMI & PUR & ACC & NMI & PUR & ACC & NMI & PUR & ACC & NMI & PUR \\ \hline
(2019) AE$^2$NET \cite{zhang2019ae2} & 0.461 & 0.320 & 0.203 & 0.514 & 0.410 & 0.312 & 0.480 & 0.388 & 0.240 & 0.676 & 0.581 & 0.352 \\
(2019) RMSL \cite{li2019reciprocal} & 0.525 & 0.474 & 0.540 & 0.554 & 0.480 & 0.554 & 0.596 & 0.551 & 0.608 & 0.354 & 0.340 & 0.391 \\
(2019) MVC-LFA \cite{wang2019multi} & 0.462 & 0.348 & 0.496 & 0.551 & 0.423 & 0.578 & 0.609 & 0.522 & 0.636 & 0.741 & 0.601 & 0.747 \\
(2019) COMIC \cite{peng2019comic} & 0.422 & 0.446 & 0.535 & 0.447 & 0.491 & 0.575 & 0.637 & 0.609 & 0.764 & 0.532 & 0.549 & 0.604 \\
(2020) CDIMC-net \cite{wen2021structural} & 0.515 & 0.480 & 0.564 & 0.528 & 0.483 & 0.565 & 0.560 & 0.564 & 0.617 & 0.727 & 0.692 & 0.742 \\
(2020) EAMC \cite{zhou2020end} & 0.419 & 0.256 & 0.427 & 0.389 & 0.214 & 0.398 & 0.356 & 0.205 & 0.370 & 0.318 & 0.173 & 0.342 \\
(2021) IMVTSC-MVI \cite{wen2021unified} & 0.490 & 0.398 & 0.540 & 0.558 & 0.445 & 0.576 & 0.687 & 0.610 & 0.719 & 0.76 & 0.691 & 0.785 \\
(2021) DEMVC \cite{xu2021deep} & 0.498 & 0.384 & 0.520 & 0.533 & 0.413 & 0.533 & 0.492 & 0.450 & 0.518 & 0.46 & 0.366 & 0.495 \\
(2021) CONAN \cite{ke2021conan}& 0.575 & 0.451 & 0.575 & 0.591 & 0.498 & 0.591 & 0.557 & 0.506 & 0.573 & 0.720 & 0.641 & 0.722 \\
(2021) CUMRL \cite{zheng2021collaborative} & 0.551 & 0.446 & 0.358 & 0.529 & 0.468 & 0.367 & 0.716 & 0.665 & 0.601 & 0.728 & 0.692 & 0.627 \\
(2021) SDMVC \cite{xu2022self}& 0.478 & 0.374 & 0.357 & 0.416 & 0.306 & 0.210 & 0.435 & 0.305 & 0.247 & 0.421 & 0.282 & 0.209 \\
(2021) CMIB \cite{wan2021multi}& 0.420 & 0.299 & 0.196 & 0.432 & 0.324 & 0.214 & 0.294 & 0.115 & 0.065 & 0.302 & 0.149 & 0.091 \\
(2021) CoMVC \cite{trosten2021reconsidering} & 0.466 & 0.426 & 0.527 & 0.541 & 0.504 & 0.584 & 0.568 & 0.569 & 0.646 & 0.700 & 0.687 & 0.746 \\
(2021) SiMVC \cite{trosten2021reconsidering} & 0.508 & 0.471 & 0.557 & 0.569 & 0.495 & 0.591 & 0.619 & 0.536 & 0.63 & 0.719 & 0.677 & 0.729 \\
(2022) DSMVC \cite{tang2022deep} & 0.603 & 0.526 & 0.619 & \underline{0.745} & \underline{0.674} & \textbf{0.745} & \underline{0.834} & \textbf{0.766} & \underline{0.834} & \underline{0.919} & \textbf{0.847} & \underline{0.919} \\
(2022) GUMRL \cite{zheng2022graph} & 0.514 & 0.508 & 0.382 & 0.700 & 0.654 & 0.551 & 0.710 & 0.667 & 0.564 & 0.686 & 0.656 & 0.539 \\
(2022) MFLVC \cite{xu2022multi} & 0.606 & \underline{0.528} & 0.616 & 0.631 & 0.566 & 0.639 & 0.733 & 0.652 & 0.734 & 0.804 & 0.703 & 0.804 \\
(2023) DealMVC \cite{yang2023dealmvc} & \underline{0.619} & 0.523 & \underline{0.619} & 0.656 & 0.589 & 0.682 & 0.757 & 0.668 & 0.757 & 0.828 & 0.732 & 0.828 \\
DCCMVC(Ours) & \textbf{0.620} & \textbf{0.546} & \textbf{0.623} & \textbf{0.790} & \textbf{0.693} & \underline{0.704} & \textbf{0.842} & \underline{0.735} & \textbf{0.836} & \textbf{0.935} & \underline{0.835} & \textbf{0.937} \\ \hline
\end{tabular}

}
\end{table*}

\begin{figure*}[htbp]
\centering
\subfigure[]{
\includegraphics[width=4.8cm]{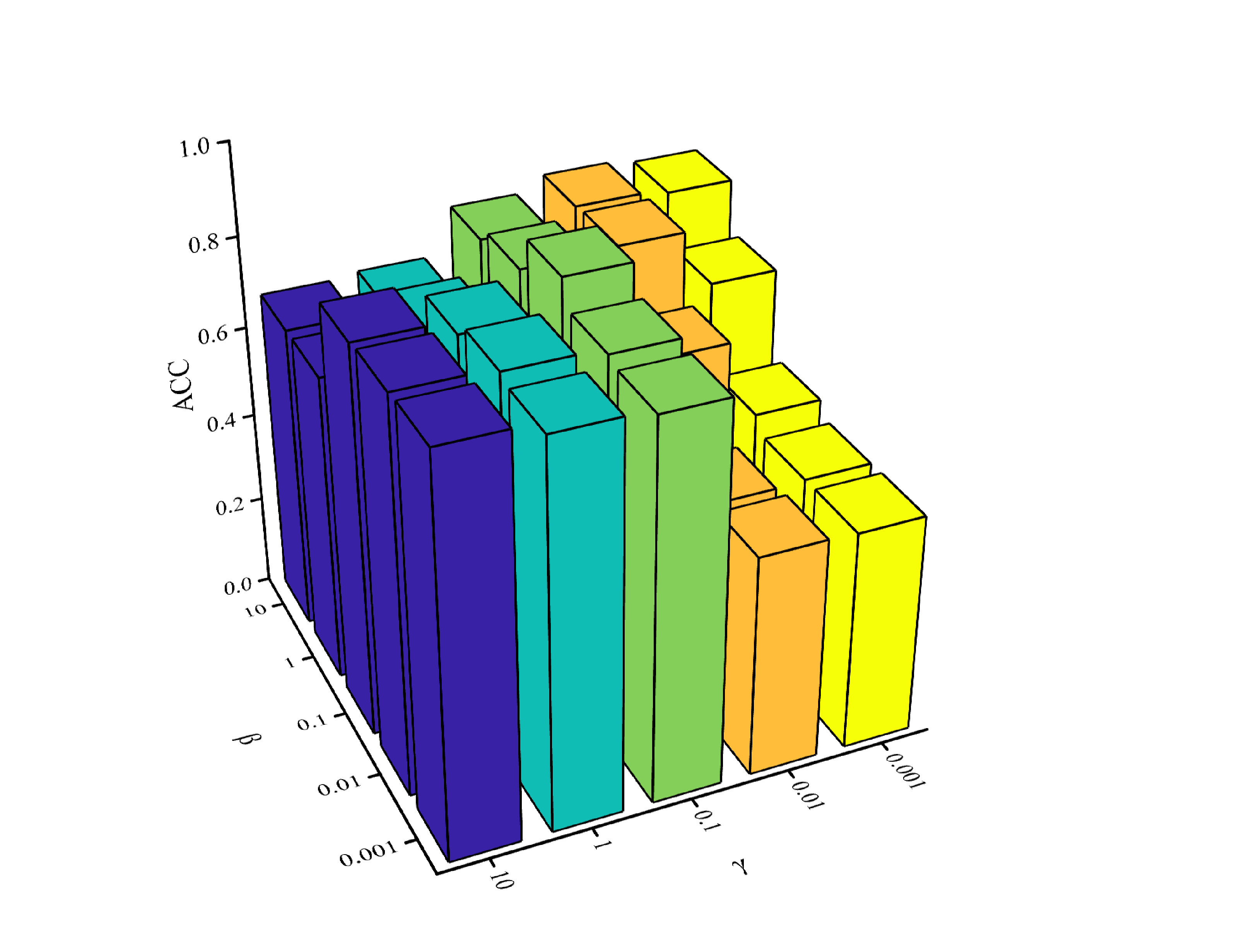}
}
\subfigure[]{
\includegraphics[width=4.8cm]{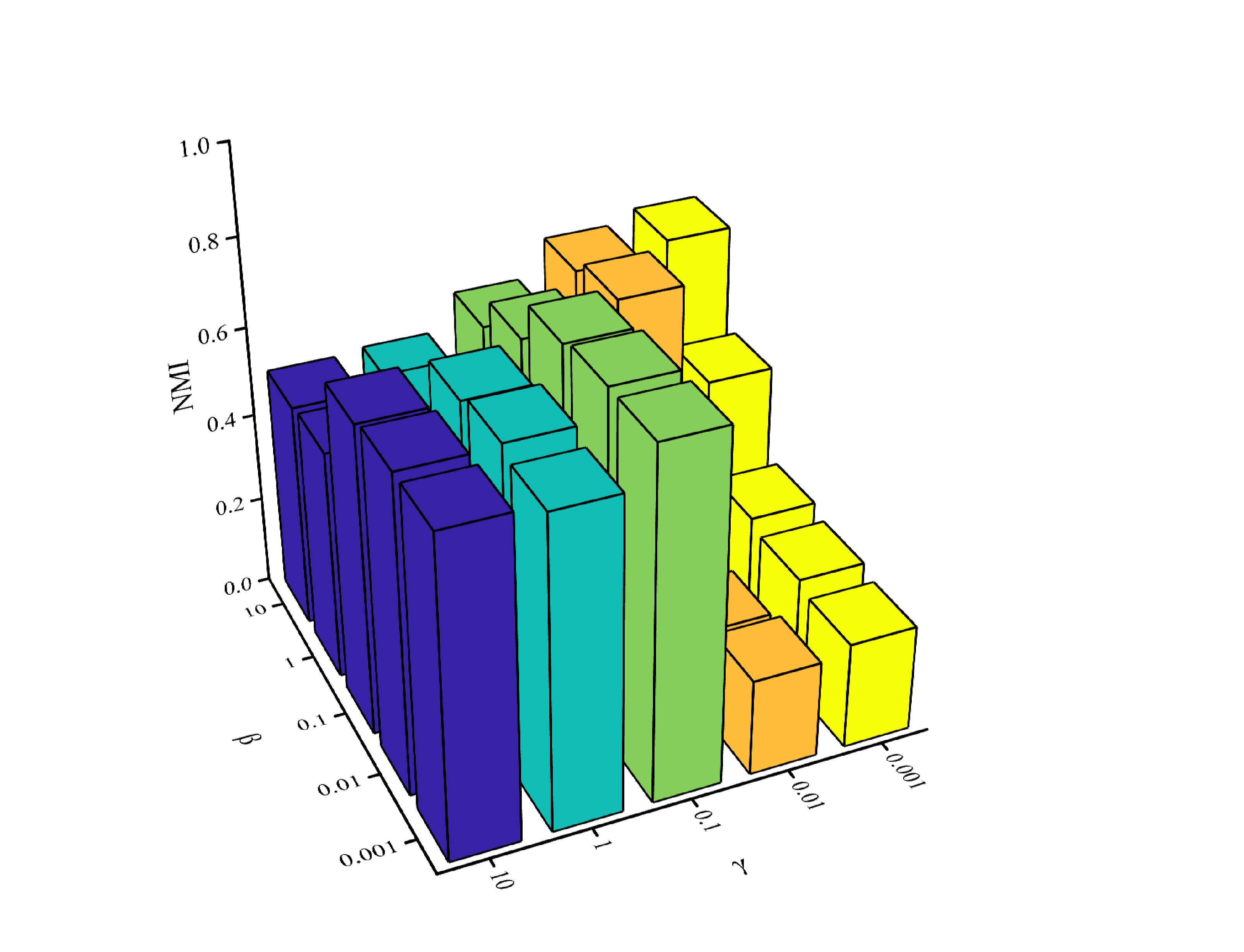}
}
\subfigure[]{
\includegraphics[width=4.8cm]{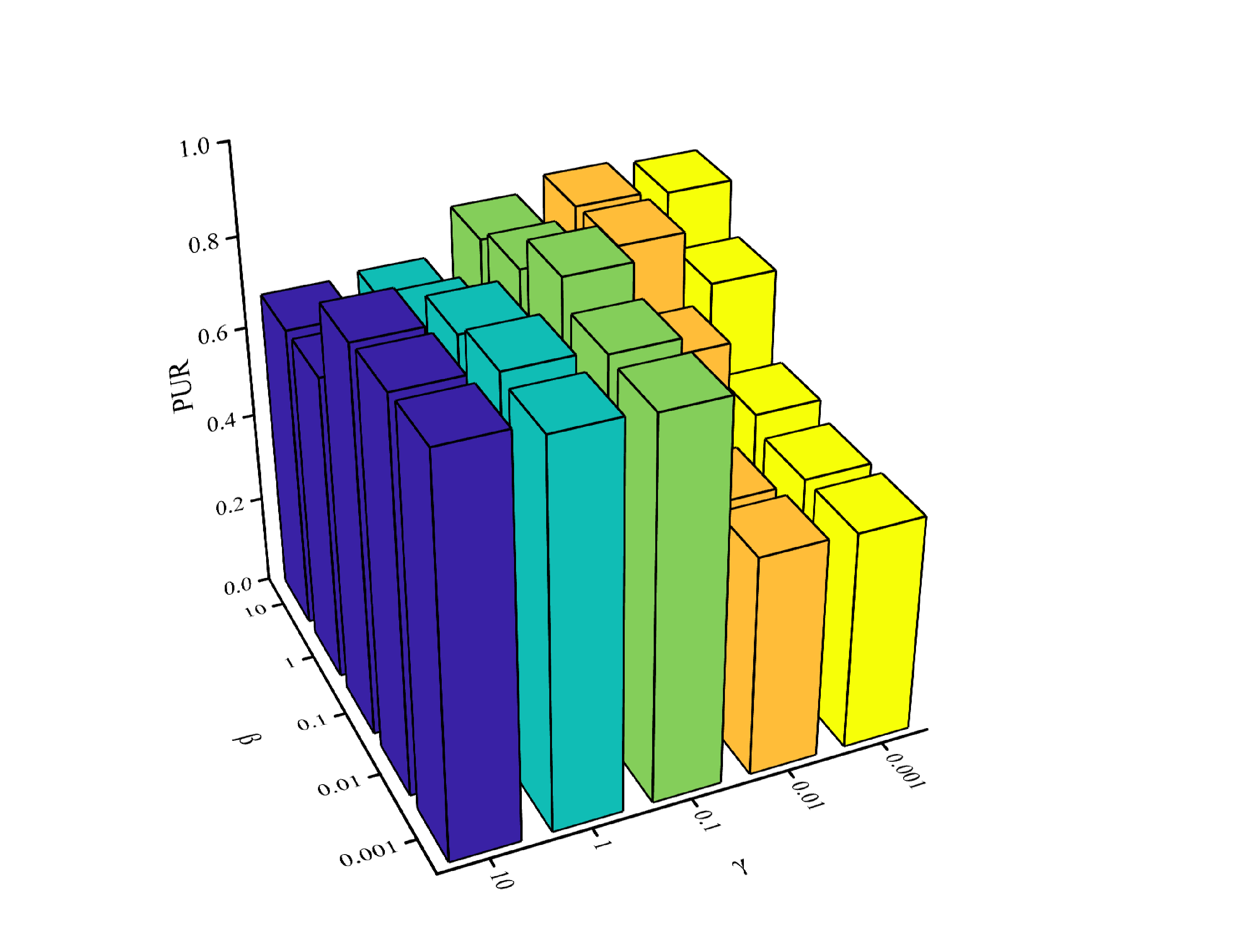}
}
\caption{Parametric analysis. The clustering performance of DCCMVC on the BBCSport dataset with different parameters $\beta$ and $\gamma$.}
\label{parametric1}
\end{figure*}

\begin{figure}[t!]
\centering
\includegraphics[width=1\linewidth]{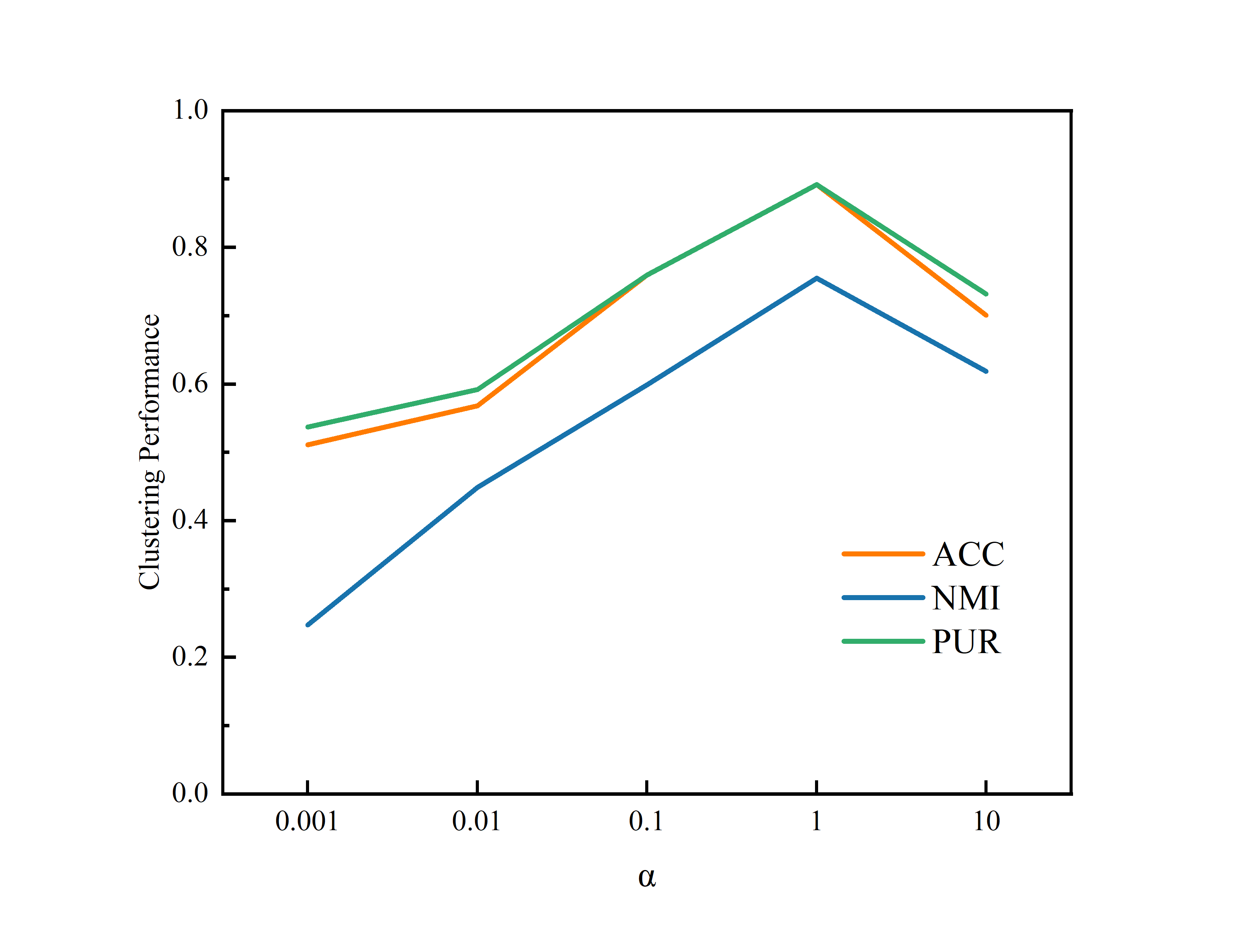}
\caption{Parametric analysis. The sensitivity analysis of the trade-off reconstruction coefficient $\alpha$ on BBCSport.}
\label{parametric2}
\end{figure}

\textbf{Evaluation metrics.}
The clustering effectiveness is evaluated by three metries, i.e., accuracy (ACC), normalized mutual information (NMI), and purity (PUR). For all these metrics, a higher value means better performance.

\textbf{Implementation details.}
The experiments were conducted on Windows with i7-12700 CPU, 16GB RAM and 3060 GPU. All datasets are reshaped into vectors, and the autoencoder is implemented using a fully connected network. We define the structure of the autoencoder as $X^{(v)}-500-500-500-2000-Z^{(v)}$, and the ReLU activation function is used to implement the variational autoencoder in DCCMVC. We first pre-trained the model for 100 epochs against reconstruction loss, then trained 100 epochs for dual consistency constraints loss, and finally fine-tuned it for 50 epochs. Adam optimizer in the Pytorch framework is adopted for optimization of the total loss with a batch size of 128 and a learning rate of 0.0001. The trade-off hyper-parameters $\alpha$, $\beta$, and $\gamma$ are set to 1, 0.01, and 0.01, respectively. For all baseline methods, we used the code published by their authors to run on our machines and use the settings recommended in their original works. The code is available at \url{https://github.com/Rumenglai/DCCMVC.git}

\begin{figure*}[htbp]
\centering
\subfigure[Convergence on BBCSport]{
\includegraphics[width=3.8cm]{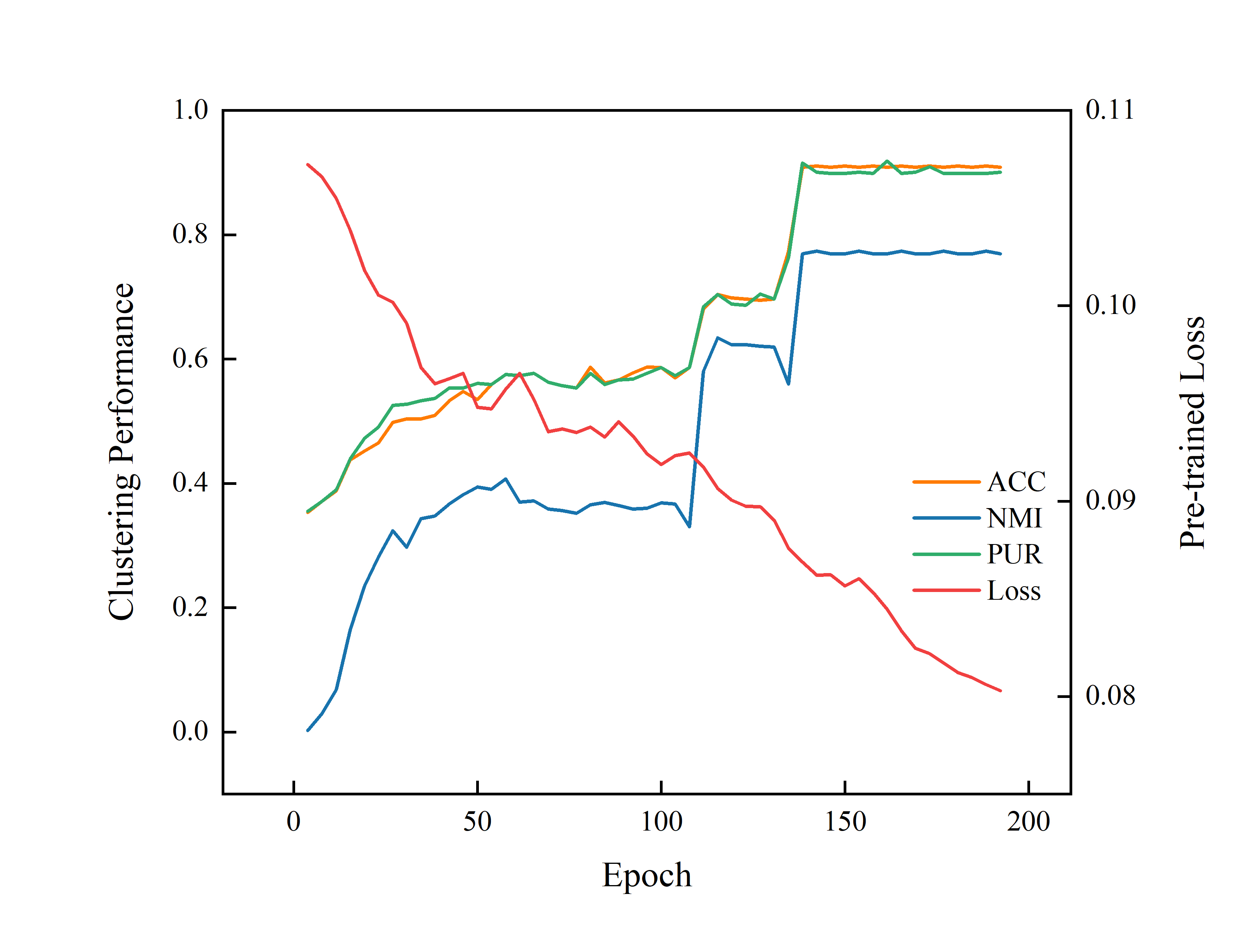}
}
\quad
\subfigure[Convergence on BBCSport]{
\includegraphics[width=3.8cm]{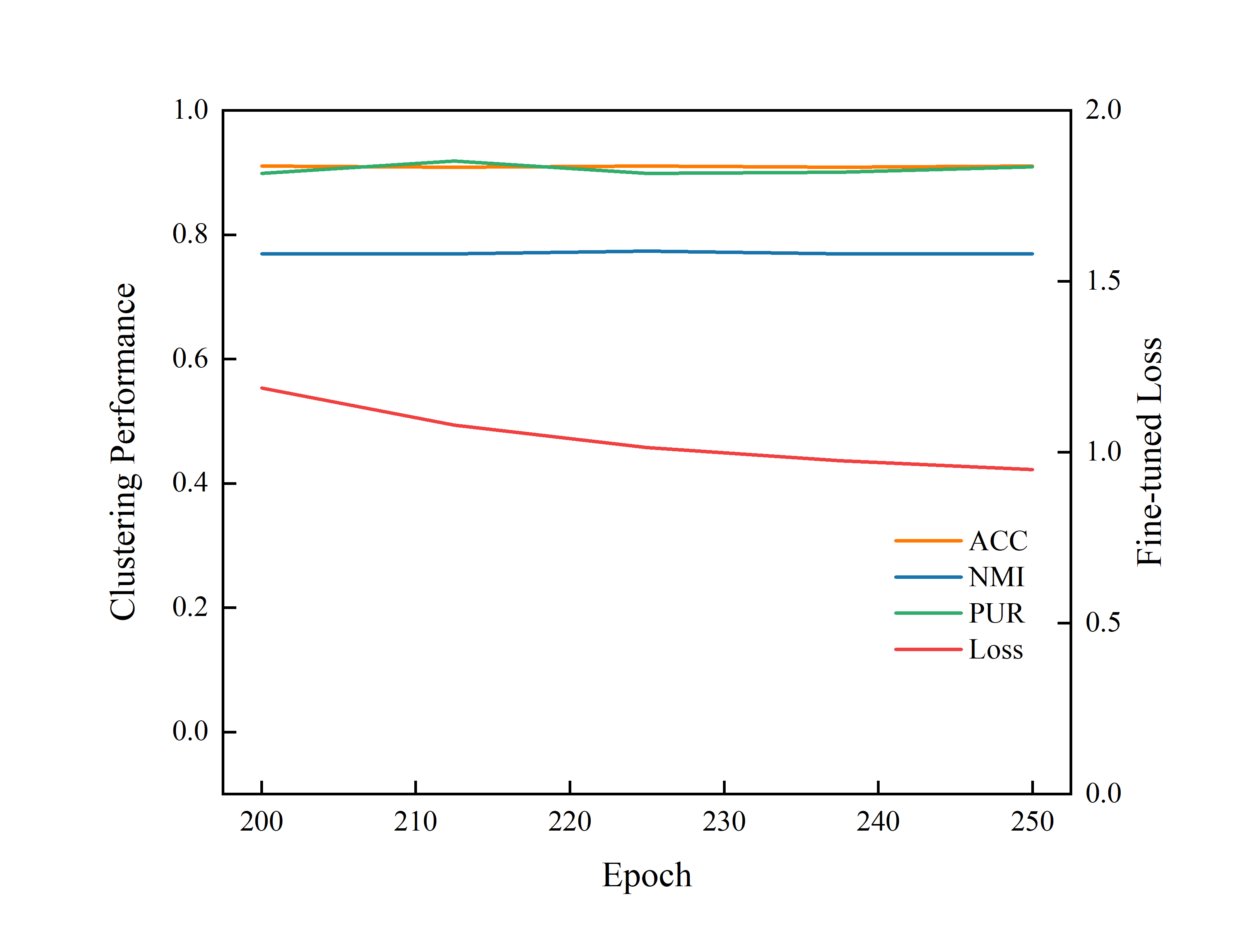}
}
\quad
\subfigure[Convergence on CCV]{
\includegraphics[width=3.8cm]{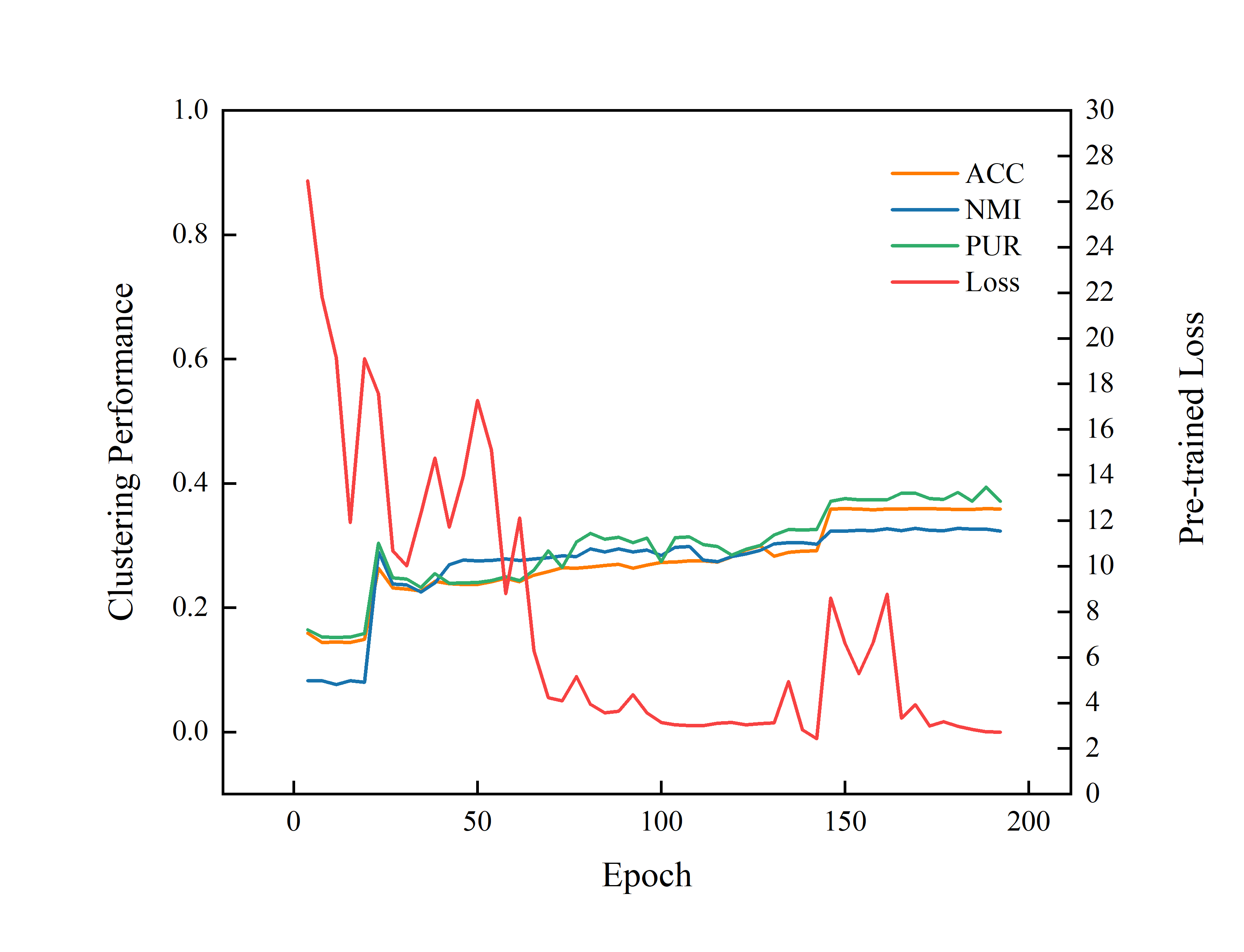}
}
\quad
\subfigure[Convergence on CCV]{
\includegraphics[width=3.8cm]{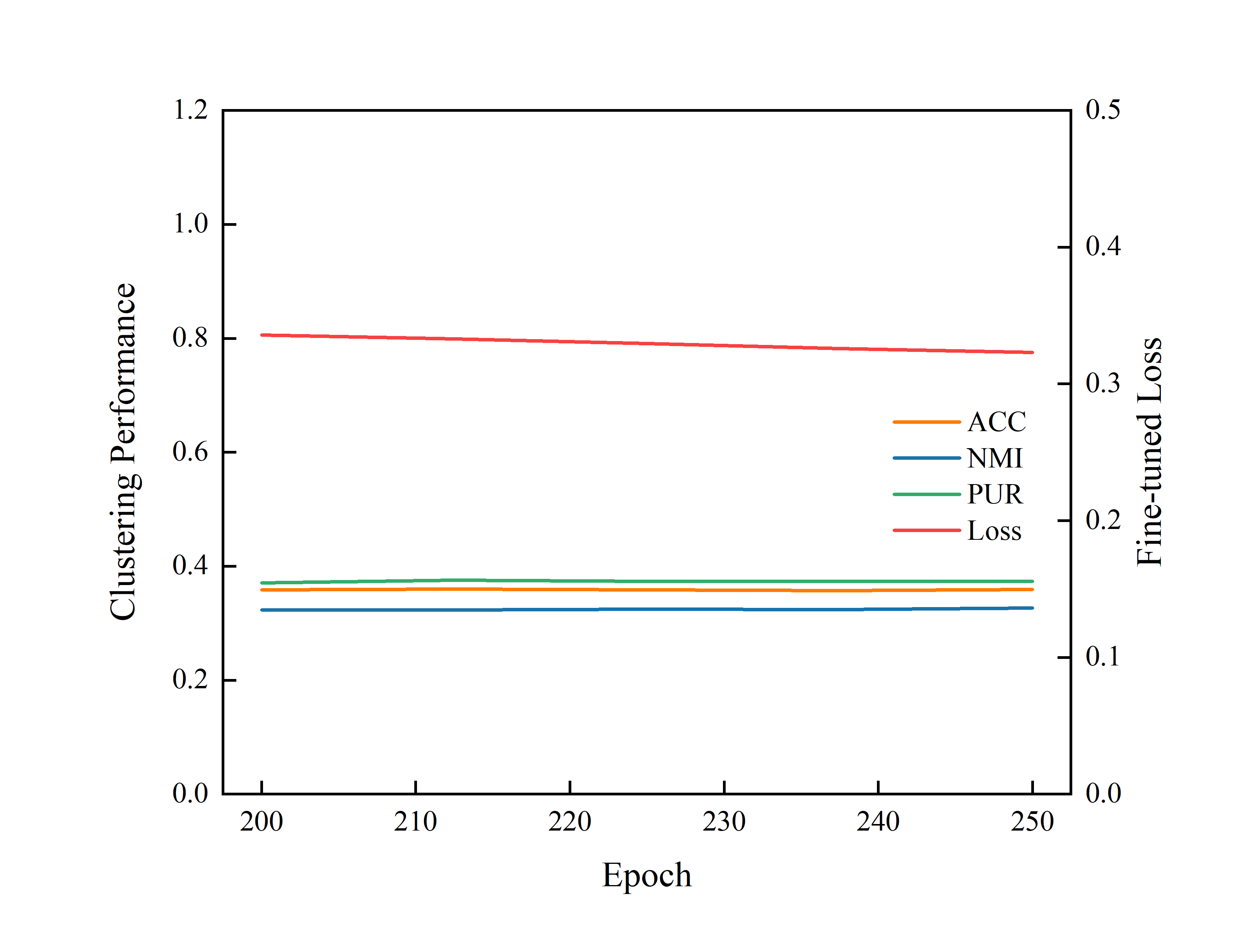}
}
\quad
\subfigure[Convergence on MNIST-USPS]{
\includegraphics[width=3.8cm]{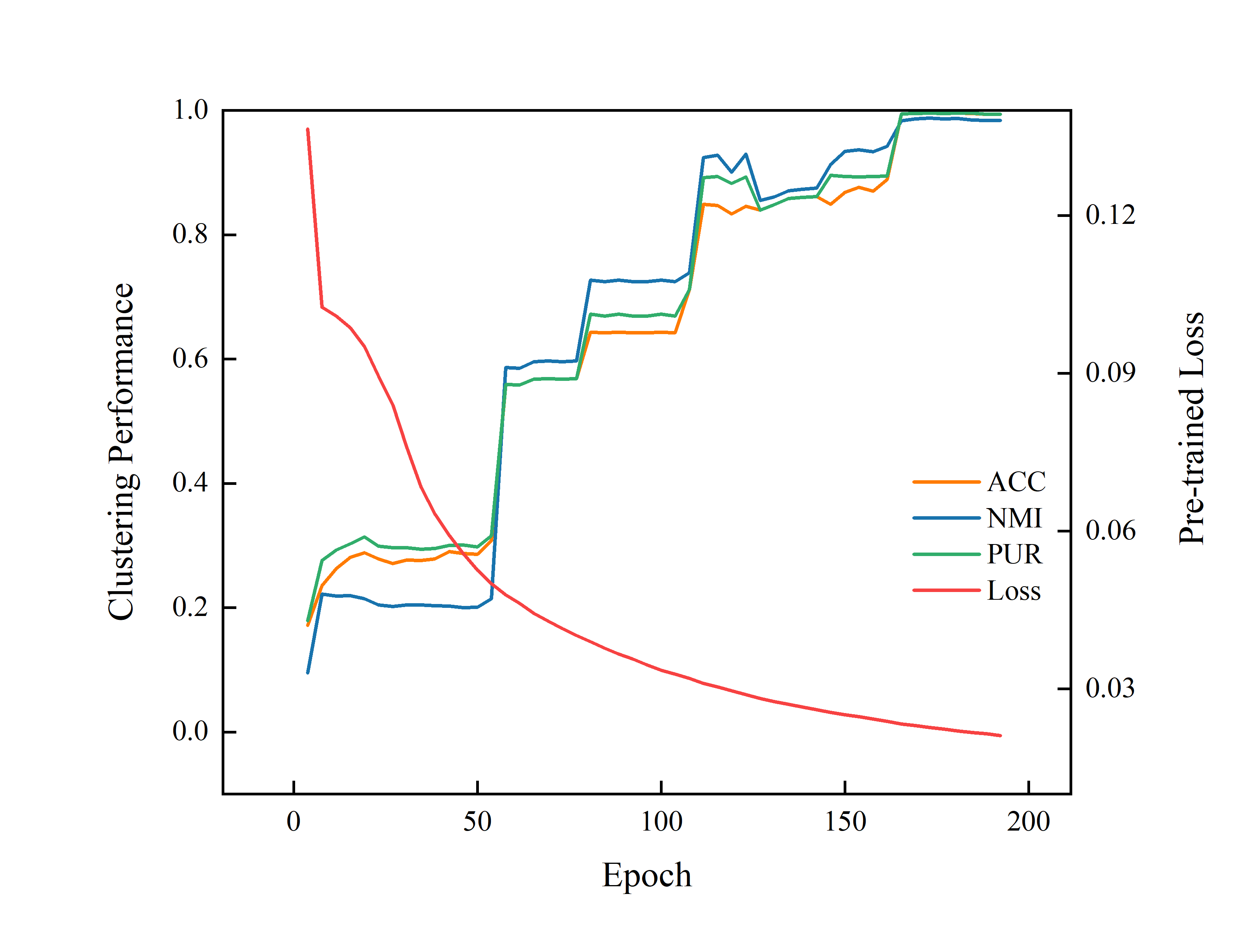}
}
\quad
\subfigure[Convergence on MNIST-USPS]{
\includegraphics[width=3.8cm]{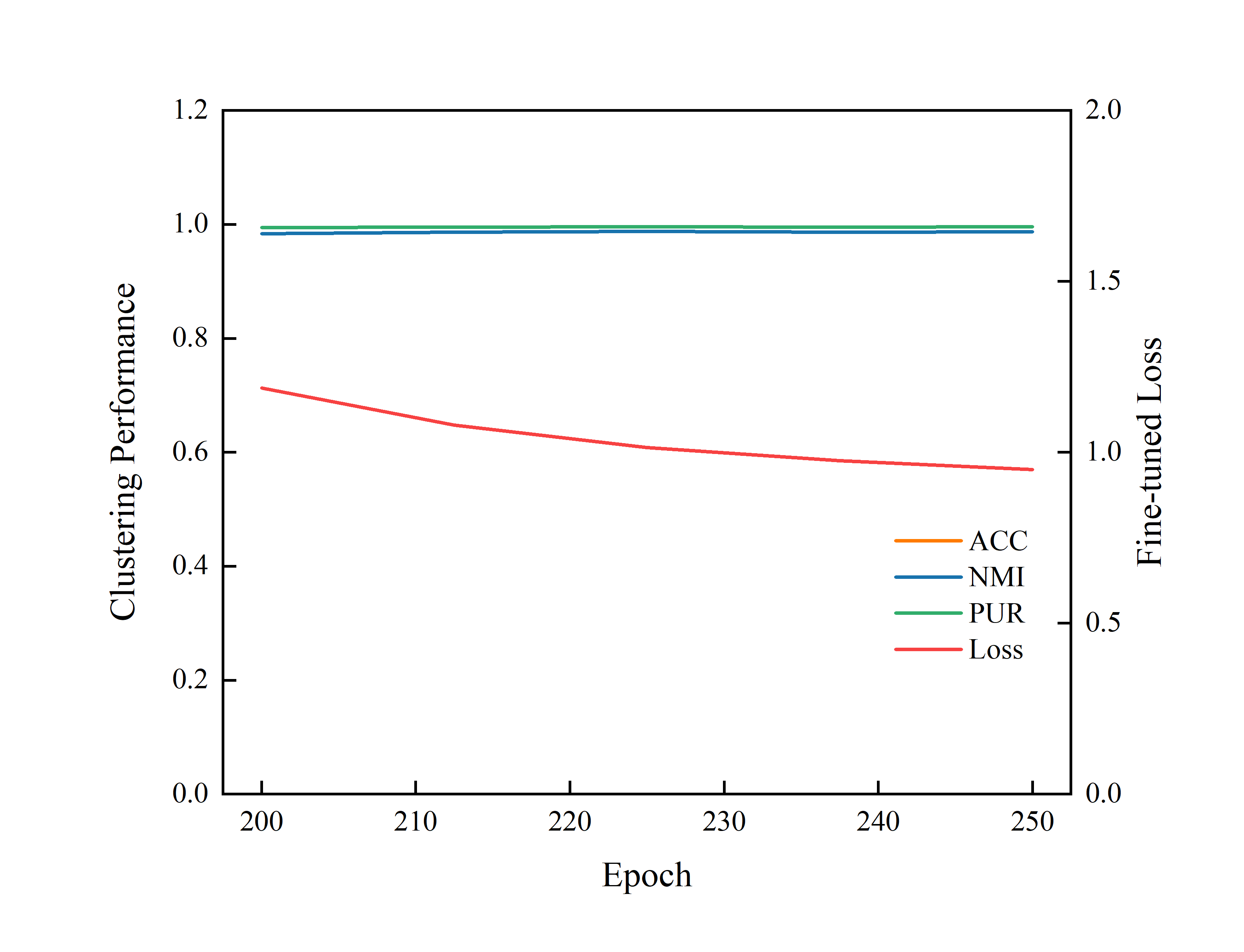}
}
\quad
\subfigure[Convergence on Reuters]{
\includegraphics[width=3.8cm]{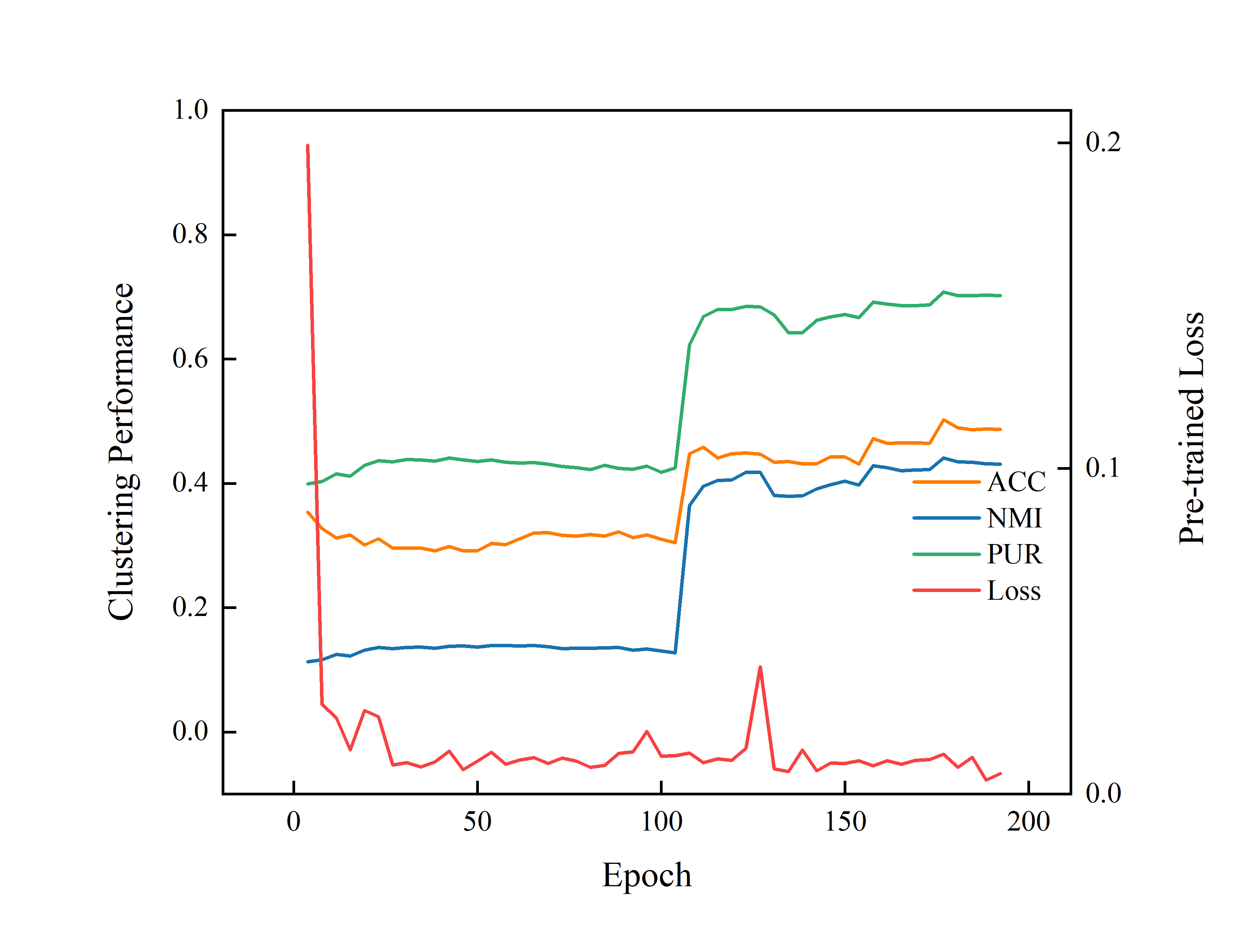}
}
\quad
\subfigure[Convergence on Reuters]{
\includegraphics[width=3.8cm]{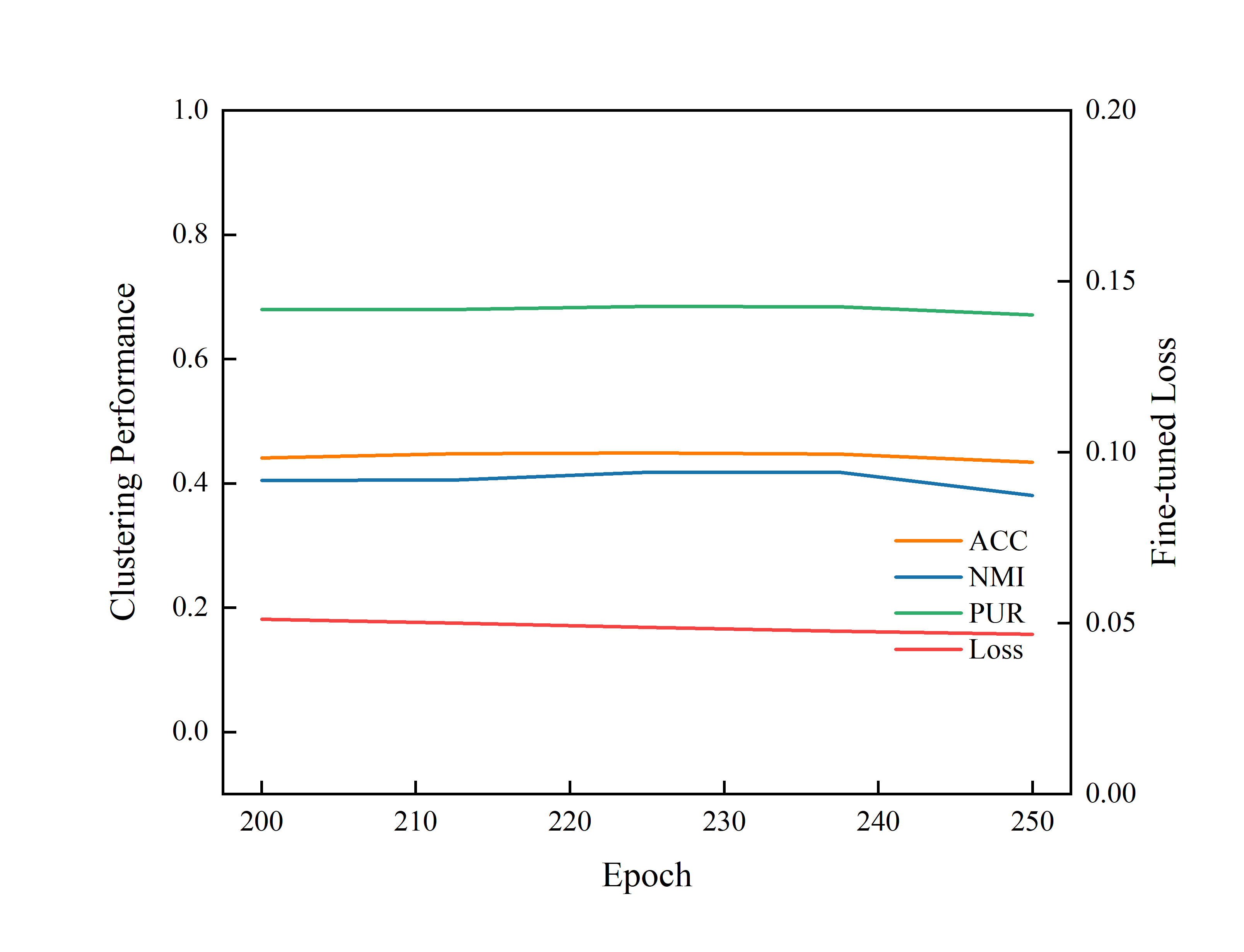}
}
\caption{Clustering performance of proposed method with increasing epoch on BBCSport (a, b), CCV (c, d), MNIST-USPS (e, f), and Reuters (g, h). The x-axis denotes the training epoch, the left and right y-axis denote the clustering performance and corresponding loss value, respectively.
}
\label{Convergence}
\end{figure*}

\subsection{Experimental Result and Analysis}
\label{sec4.2}
The clustering performance of all methods on 8 datasets are presented as Table. \ref{table2} and Table. \ref{table3}. We obtain the following observations: i) The proposed method is robust on different datasets, which achieves the best or second best performance on eight datasets. Compared the second best method, the proposed method has a significant improvement, especially surpassing 8.2\% on the BBCSport dataset. Moreover, DCCMVC achieves 14.1\% and 27.8\% performance improvements over the second best baseline on Reuters in terms of NMI and PUR. ii) In datasets with more than two views such as CCV, Caltech-3V, 4V, and 5V, the proposed method reduces the information that affects the clustering effect by disentangling the shared and private information between views, and constraining the consistency of the shared information. The data of each different view of the Caltech dataset not only contains the shared information of samples but also has a large amount of private information and noisy information, which information increases with the number of views. Thus, the introduction of contrastive learning improves the learning of multi-view data consistency information. Since our method learns the public representation of multi-view, more view data is more instructive for clustering, and the proposed method is also suitable for multi-view tasks. iii) When examining the outcomes on eight datasets, the proposed method significantly incrementally outperforms the AE$^2$NET, COMIC, DEMVC, SDMVC, DSMVC, and CoMVC methods on ACC, NMI, and PUR by significant increments. We conjecture the reason is that those methods do not take into account the contrastive learning mechanism.

\subsection{Parameter Sensitivity Analysis}
\label{sec4.3}
To evaluate the robustness of hyper-parameters on the clustering performance of proposed methed, which includes the trade-off reconstruction coefficient $\alpha$, shared information consistency coefficient $\beta$ and cross-view consistency coefficient $\gamma$. We conduct experiments using the BBCSport. As shown in Fig. \ref{parametric1} and Fig. \ref{parametric2}, which illustrate that our model is sensitive to all parameters. Therefore, we fix $\alpha$ to 0.001, 0.01, 0.1, 1 and 10 respectively to found best one, and eventual $\alpha$ is set to 1 here. Empirically, we set the remaining parameters $\beta$ and $\gamma$ to 0.1 and 0.1, which achieves the best results, i.e., $\alpha$ is set to 1,  $\beta$ is set to 0.01, and $\gamma$ is set to 0.01.

\subsection{Convergence Analysis}
\label{sec4.4}
To investigate the convergence, we report the loss values and the corresponding clustering performance of the proposed method through iterations on 4 datasets, BBCSport, CCV, MNIST-USPS, and Reuters, in Fig. \ref{Convergence}. One could observe that on the 4 datasets, the loss value of the pre-trained process monotonically decreases and eventually levels off and remains consistent over a small range until convergence, and various evaluation metrics continuously increase and tend to be smoothed and consistent.

\begin{table*}[ht!]
\caption{Ablation studies of the proposed method. 'R', 'S', 'P', 'SI', and 'CL' denote reconstruction, shared information, private information, shared information consistency inference and cross-view consistency maximization, respectively. 
}
\label{table4}
\renewcommand\arraystretch{1.5}
\centering
\footnotesize
\setlength{\tabcolsep}{2.8mm}{
\begin{tabular}{lccccccc}
\hline
\multirow{2}{*}{Datasets} & \multirow{2}{*}{Metrics} & \multicolumn{6}{c}{Components} \\ \cline{3-8} 
 &  & (1) R & (2) R\&S & (3) R\&S\&SI & (4) R\&S\&P & (5) R\&S\&P\&SI & (6) R\&S\&P\&SI\&CL(\textbf{Ours}) \\ \hline
\multirow{3}{*}{BBCSport} & ACC & 0.356 & 0.435 & 0.602 & 0.727 & 0.829 & 0.889 \\
 & NMI & 0.088 & 0.205 & 0.447 & 0.601 & 0.775 & 0.732 \\
 & PUR & 0.428 & 0.494 & 0.606 & 0.729 & 0.832 & 0.843 \\ \hline
\multirow{3}{*}{CCV} & ACC & 0.184 & 0.211 & 0.246 & 0.251 & 0.287 & 0.359 \\
 & NMI & 0.187 & 0.229 & 0.277 & 0.274 & 0.291 & 0.323 \\
 & PUR & 0.201 & 0.214 & 0.248 & 0.251 & 0.302 & 0.371 \\ \hline
\multirow{3}{*}{MNIST-USPS} & ACC & 0.484 & 0.550 & 0.793 & 0.863 & 0.886 & 0.995 \\
 & NMI & 0.713 & 0.609 & 0.893 & 0.873 & 0.926 & 0.993 \\
 & PUR & 0.484 & 0.550 & 0.794 & 0.863 & 0.886 & 0.995 \\ \hline
\multirow{3}{*}{Reuters} & ACC & 0.211 & 0.346 & 0.410 & 0.412 & 0.413 & 0.465 \\
 & NMI & 0.346 & 0.328 & 0.404 & 0.354 & 0.375 & 0.404 \\
 & PUR & 0.621 & 0.619 & 0.678 & 0.651 & 0.643 & 0.761 \\ \hline
\end{tabular}
}
\end{table*}

\begin{figure*}[ht!]
\centering
\subfigure[\scriptsize{Epoch 10(ACC=0.371)}]{
\includegraphics[width=2.8cm]{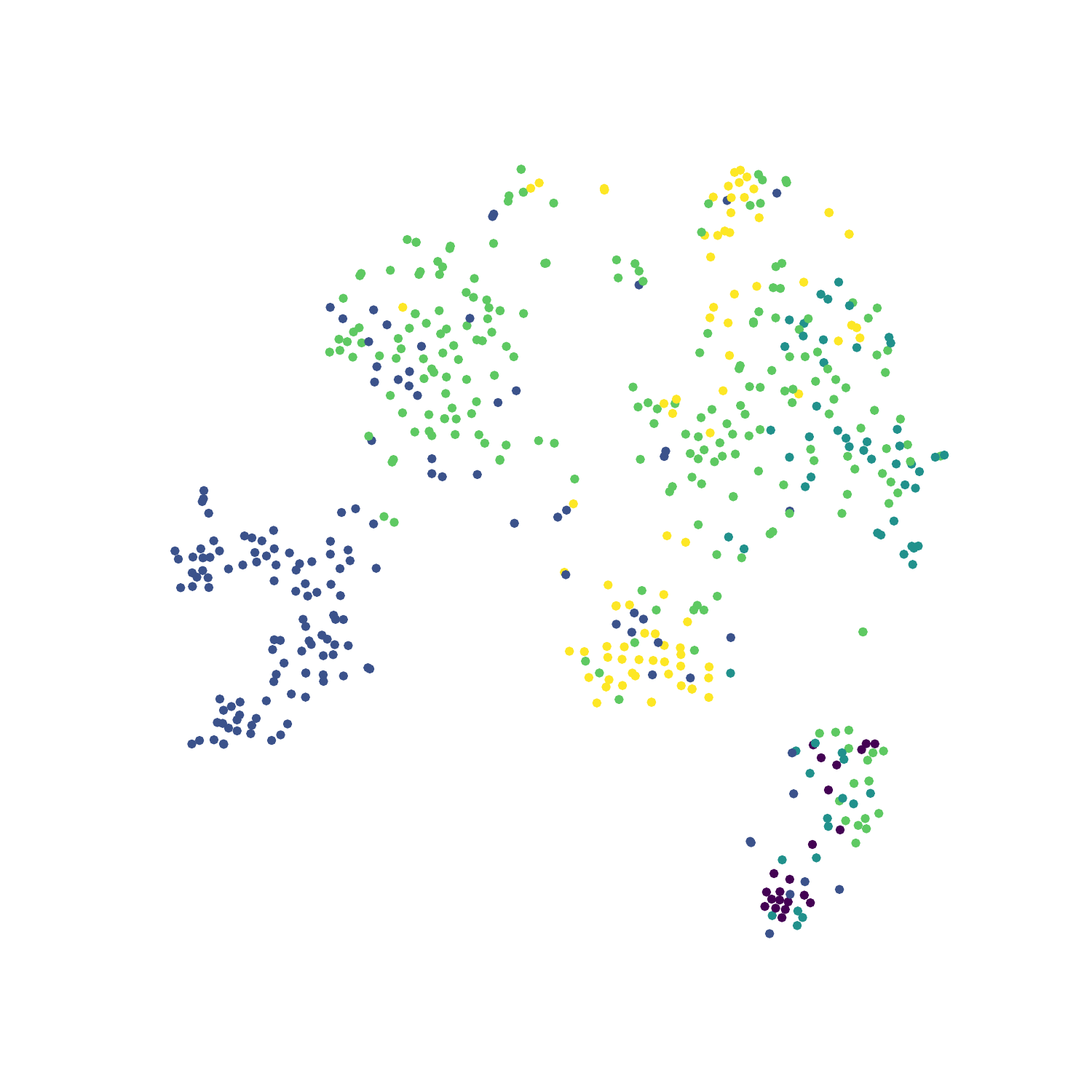}
}
\quad
\subfigure[\scriptsize{Epoch 50(ACC=0.549)}]{
\includegraphics[width=2.8cm]{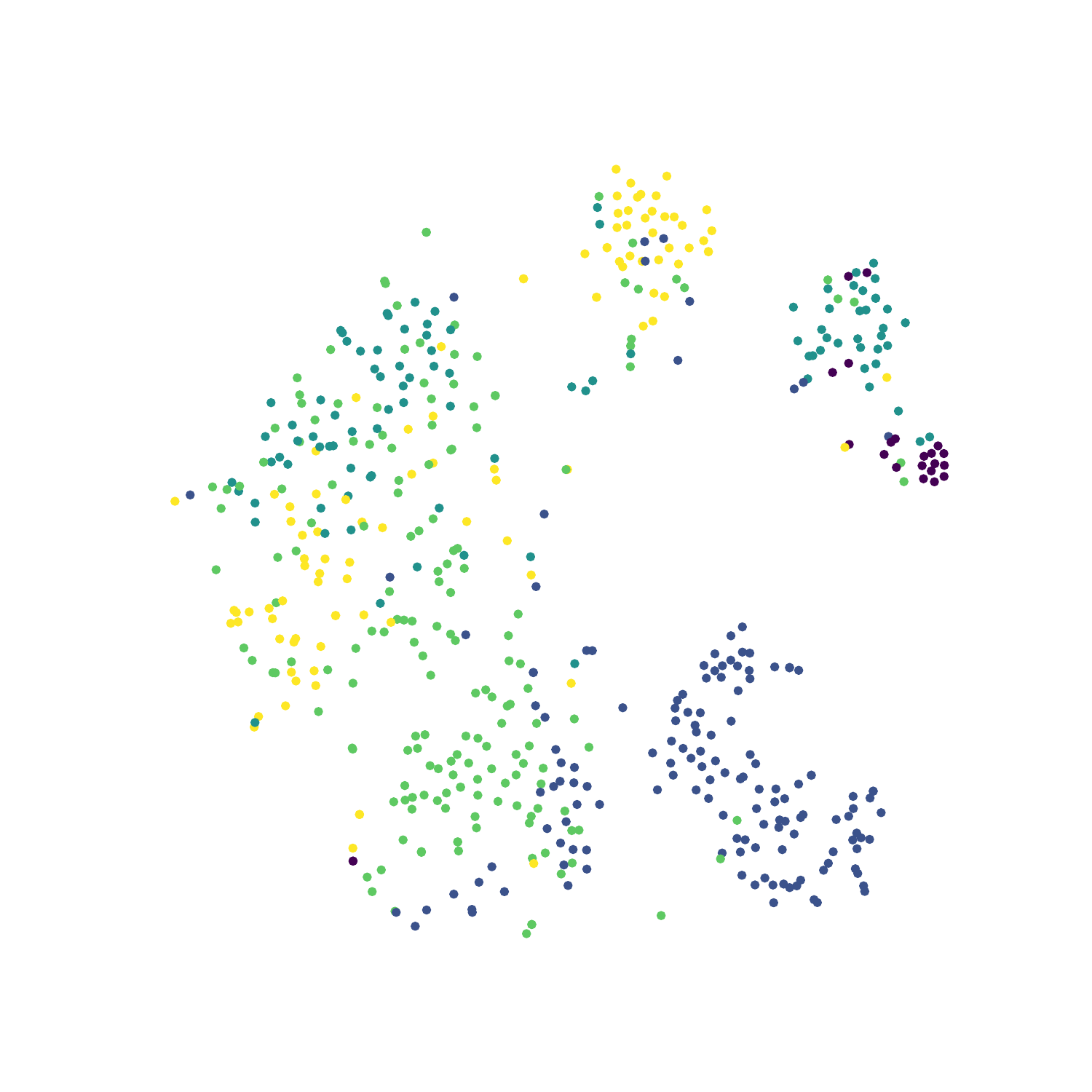}
}
\quad
\subfigure[\scriptsize{Epoch 100(ACC=0.553)}]{
\includegraphics[width=2.8cm]{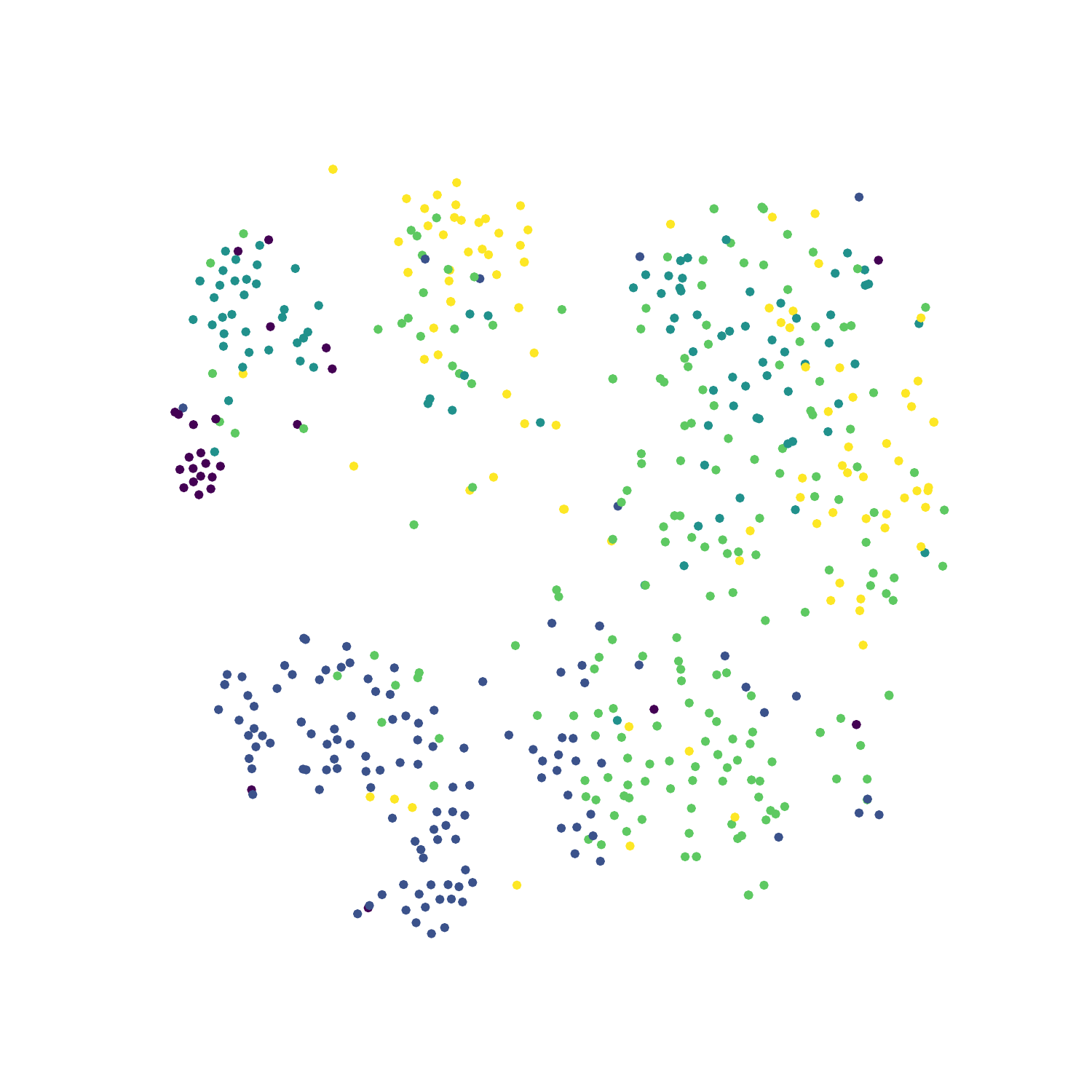}
}
\quad
\subfigure[\scriptsize{Epoch 200(ACC=885)}]{
\includegraphics[width=2.8cm]{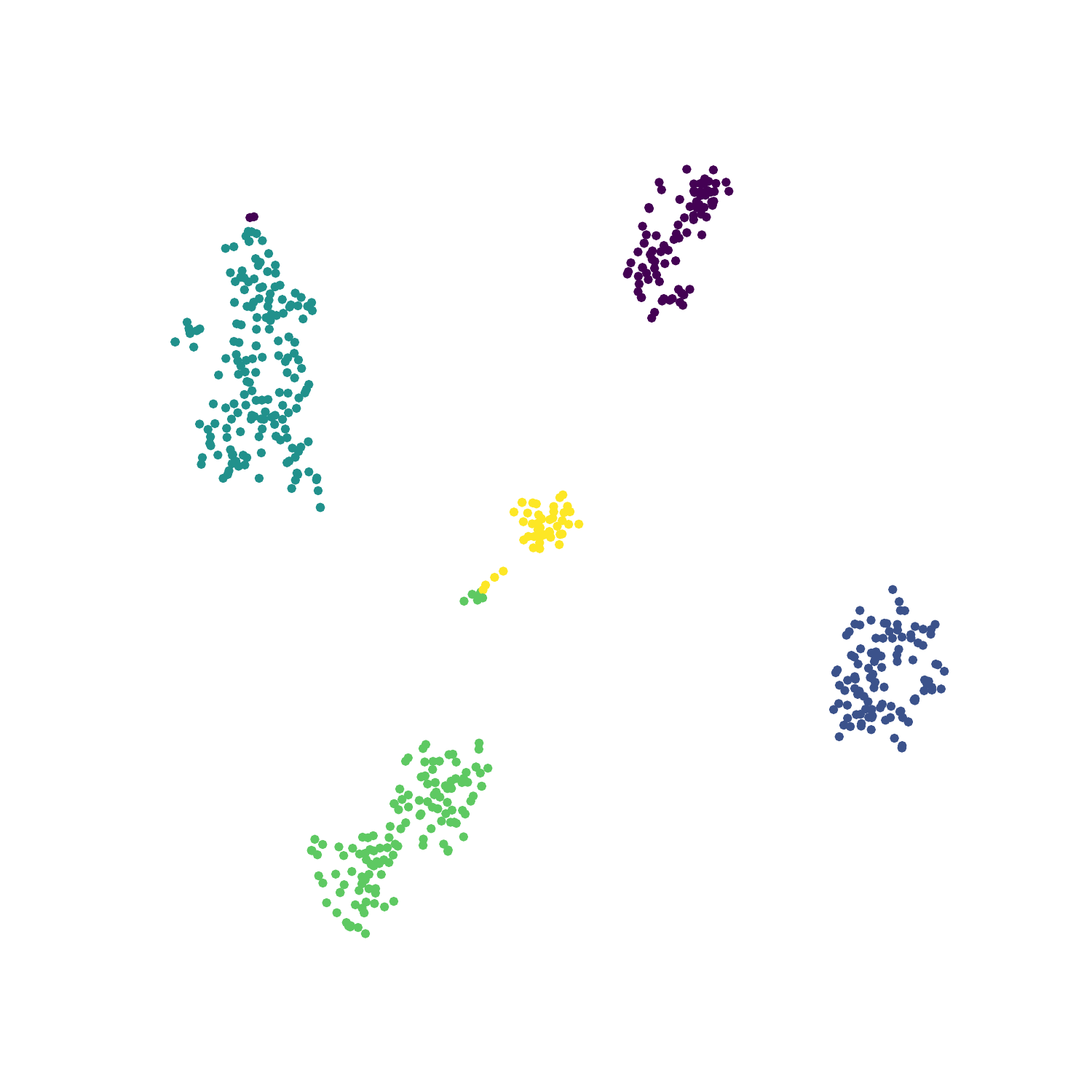}
}
\quad
\subfigure[\scriptsize{Epoch 250(ACC=0.889)}]{
\includegraphics[width=2.8cm]{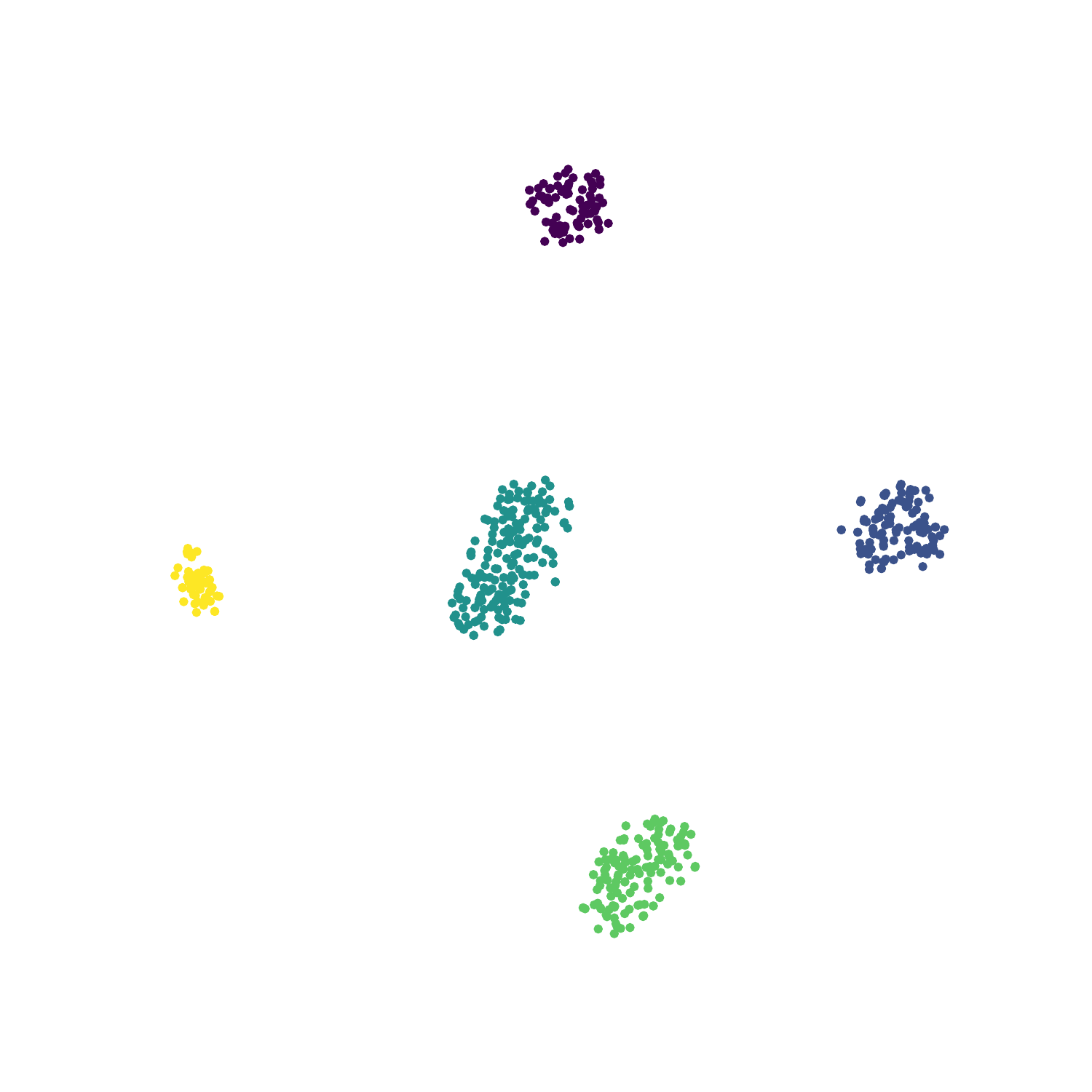}
}
\quad
\subfigure[\scriptsize{Epoch 10(ACC=0.158)}]{
\includegraphics[width=2.8cm]{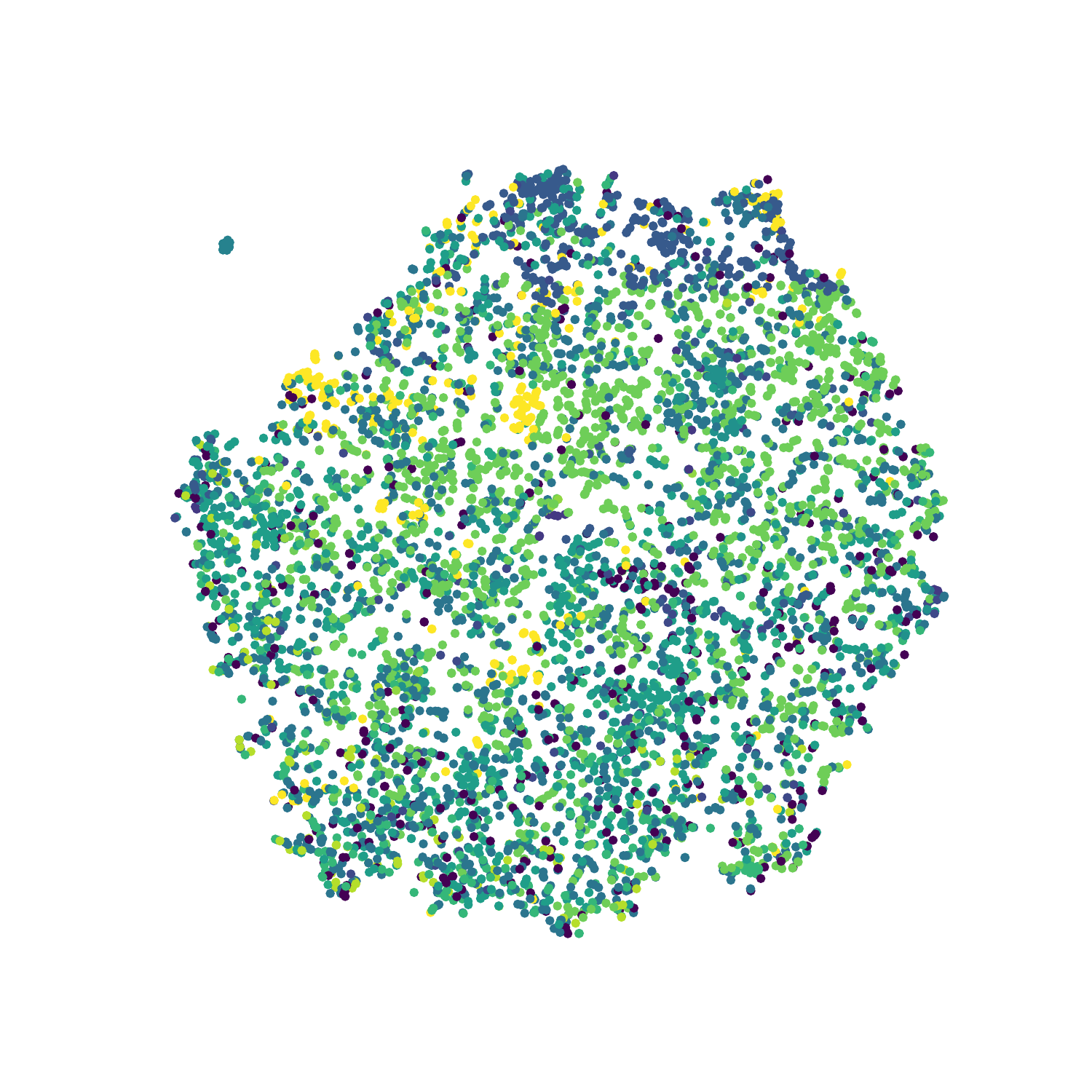}
}
\quad
\subfigure[\scriptsize{Epoch 50(ACC=0.146)}]{
\includegraphics[width=2.8cm]{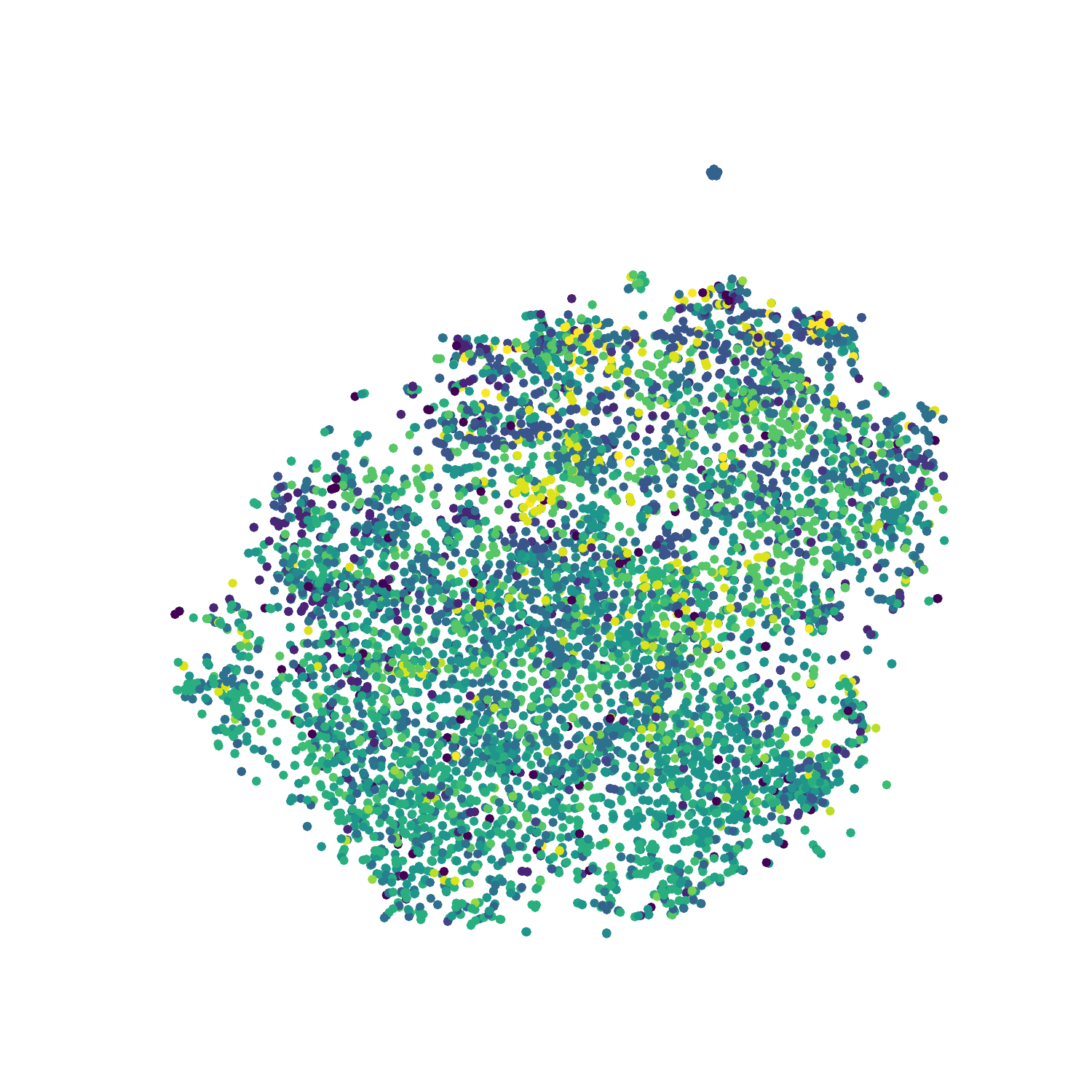}
}
\quad
\subfigure[\scriptsize{Epoch 100(ACC=0.158)}]{
\includegraphics[width=2.8cm]{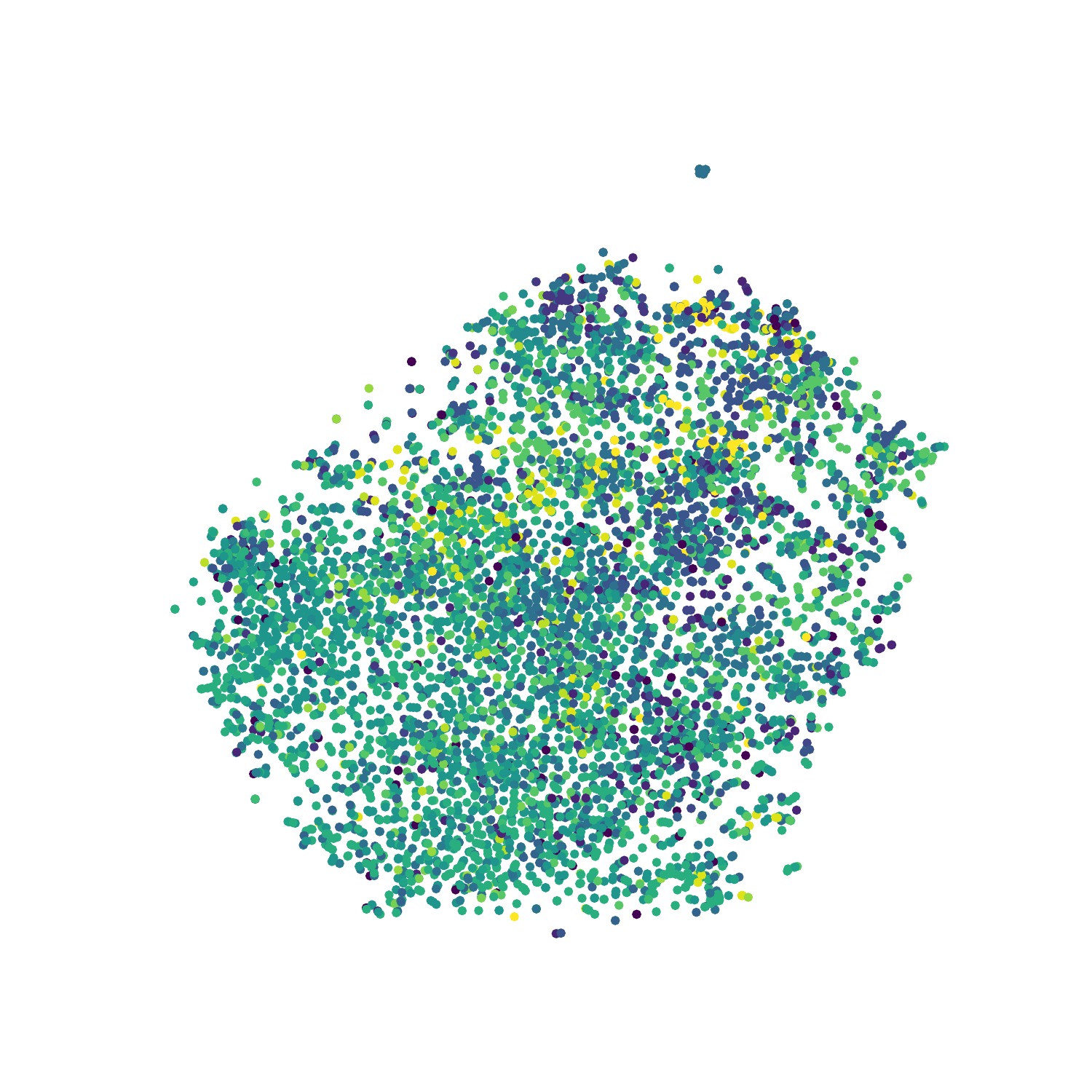}
}
\quad
\subfigure[\scriptsize{Epoch 200(ACC=0.262)}]{
\includegraphics[width=2.8cm]{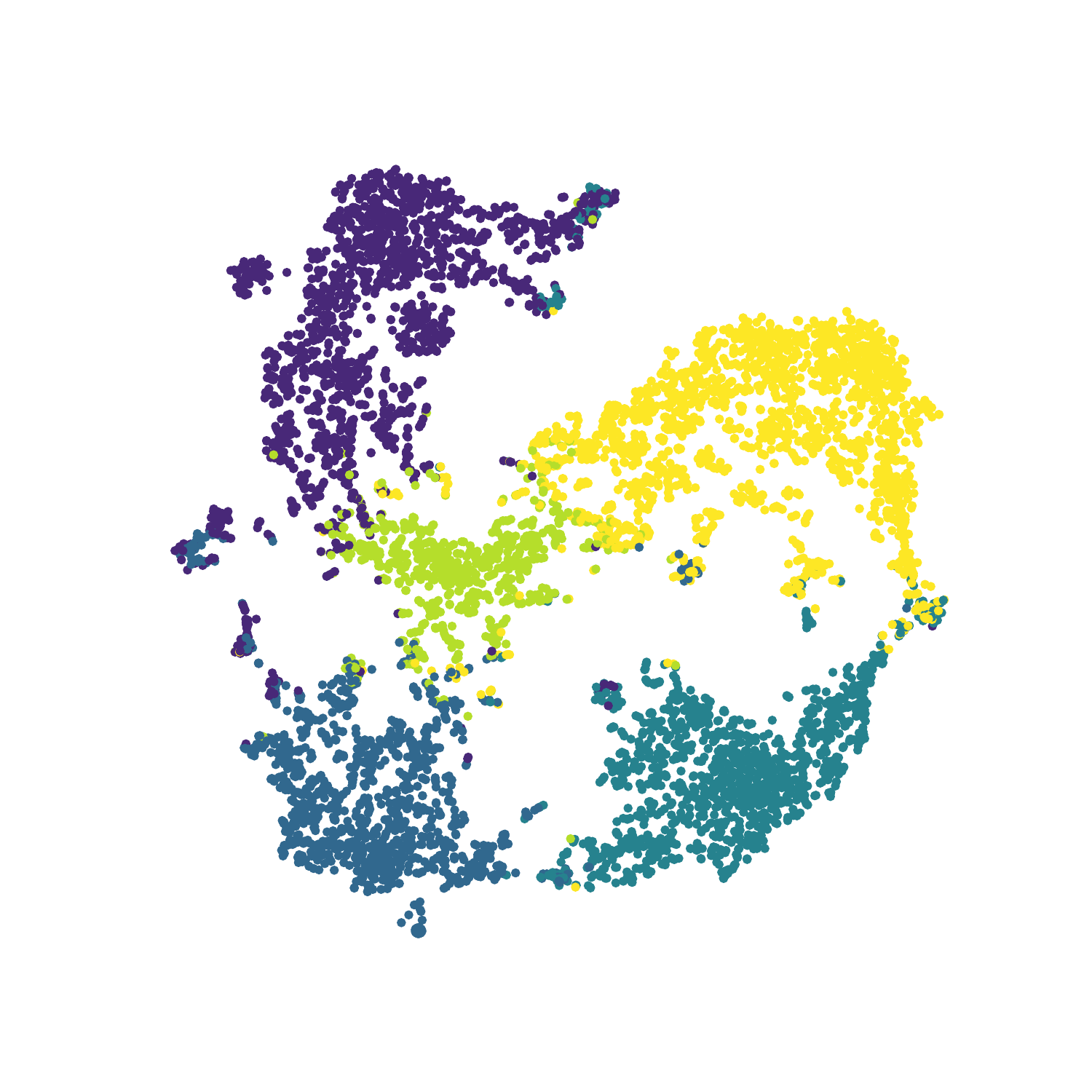}
}
\quad
\subfigure[\scriptsize{Epoch 250(ACC=0.359)}]{
\includegraphics[width=2.8cm]{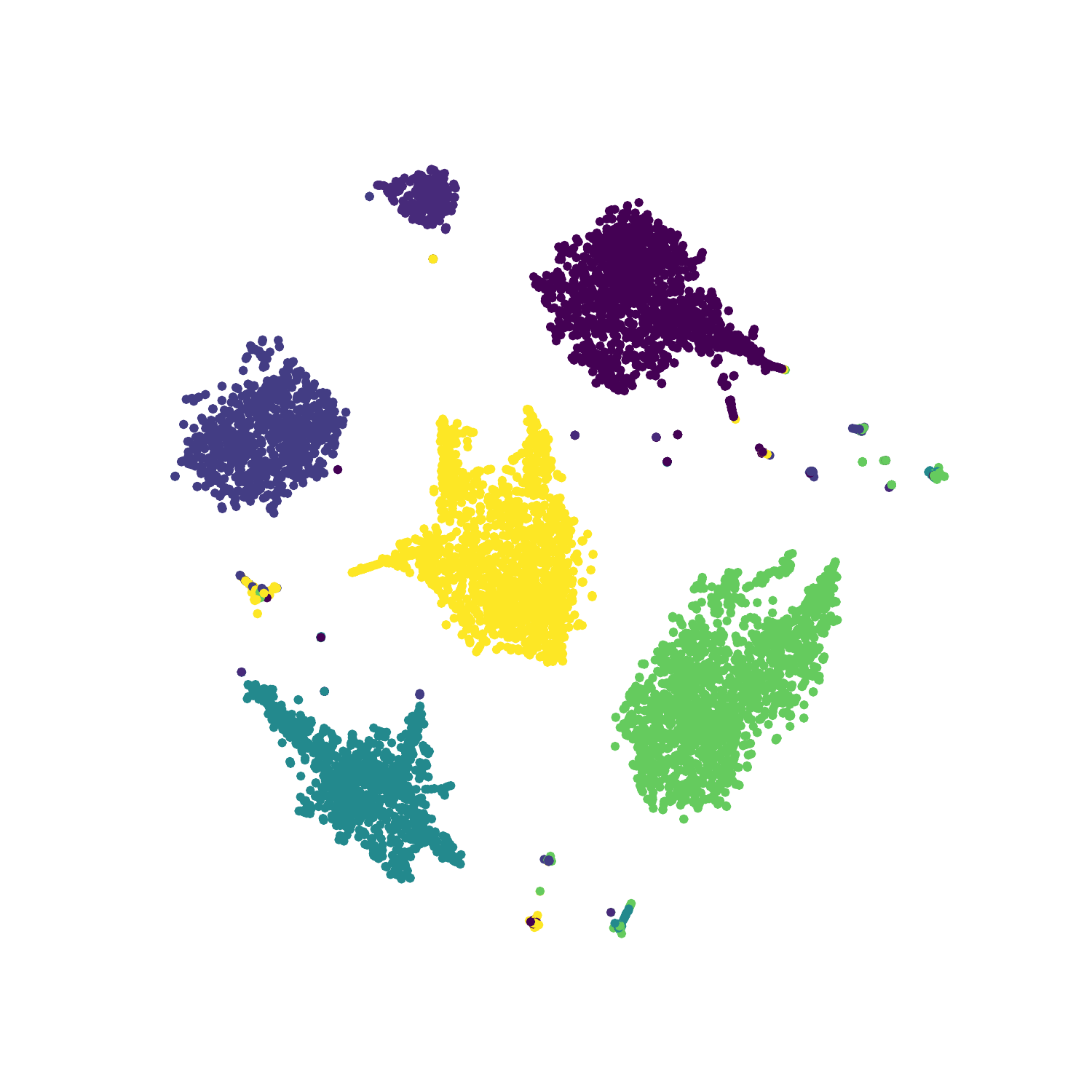}
}
\quad
\subfigure[\scriptsize{Epoch 10(ACC=0.145)}]{
\includegraphics[width=2.8cm]{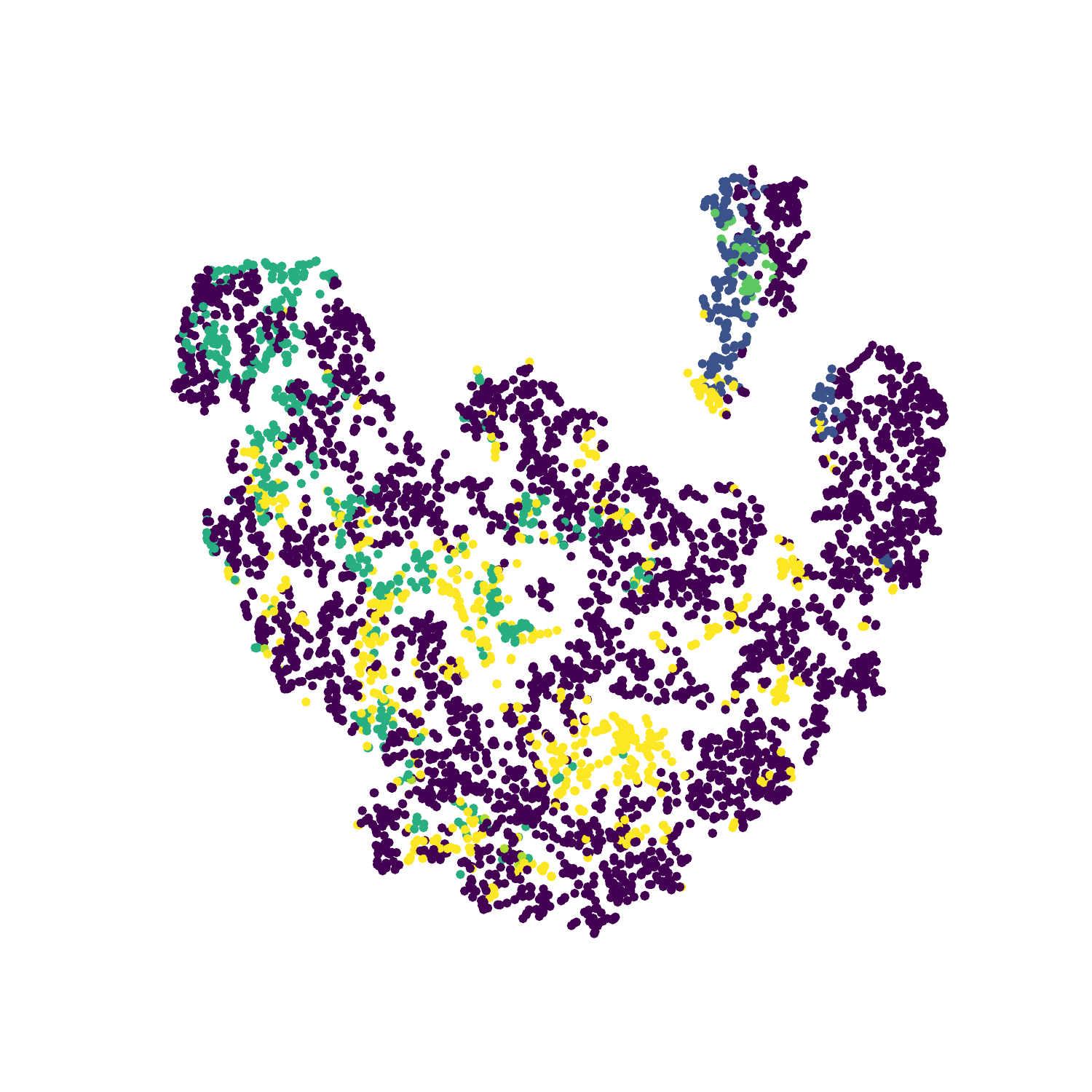}
}
\quad
\subfigure[\scriptsize{Epoch 50(ACC=0.249)}]{
\includegraphics[width=2.8cm]{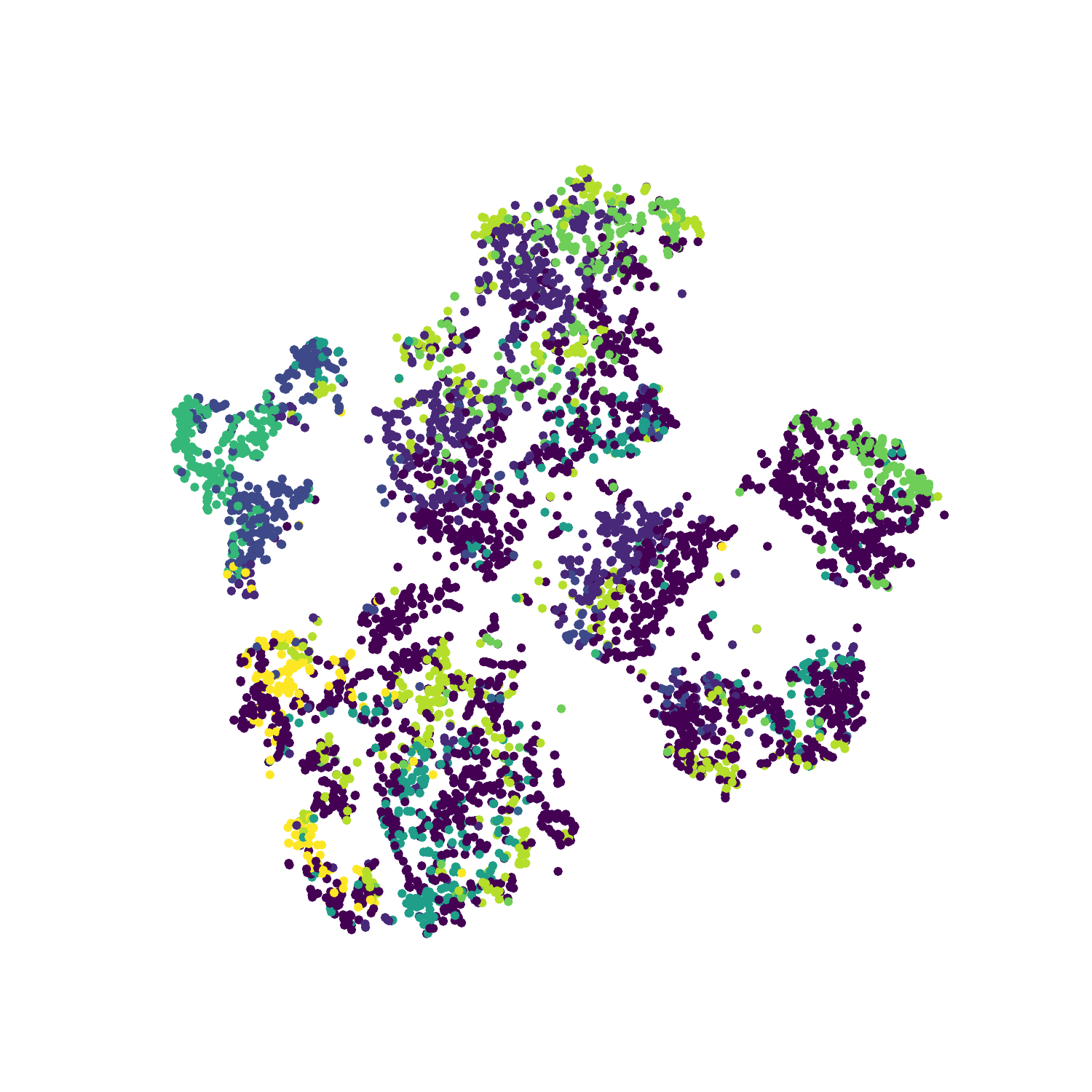}
}
\quad
\subfigure[\scriptsize{Epoch 100(ACC=0.853)}]{
\includegraphics[width=2.8cm]{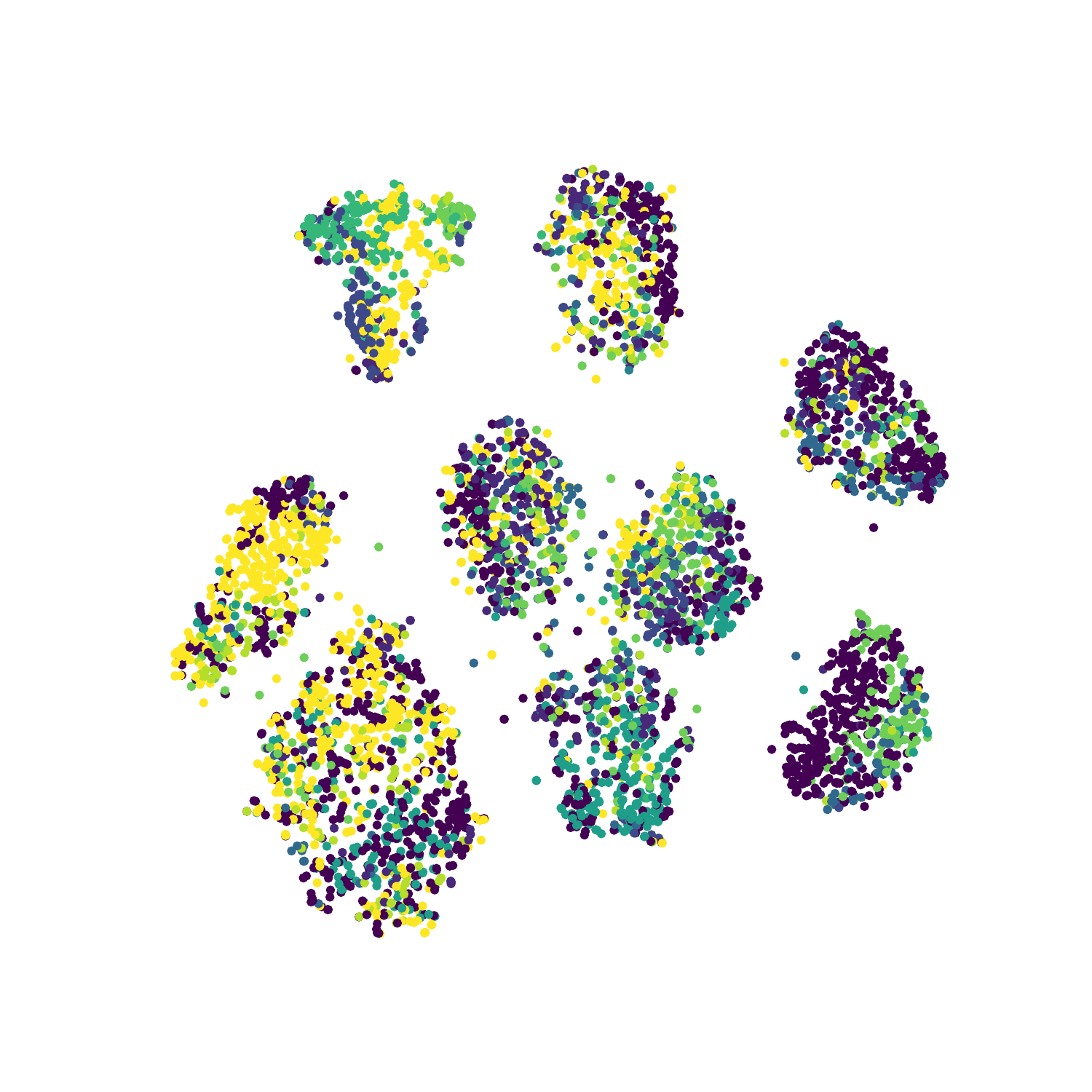}
}
\quad
\subfigure[\scriptsize{Epoch 200(ACC=0.993)}]{
\includegraphics[width=2.8cm]{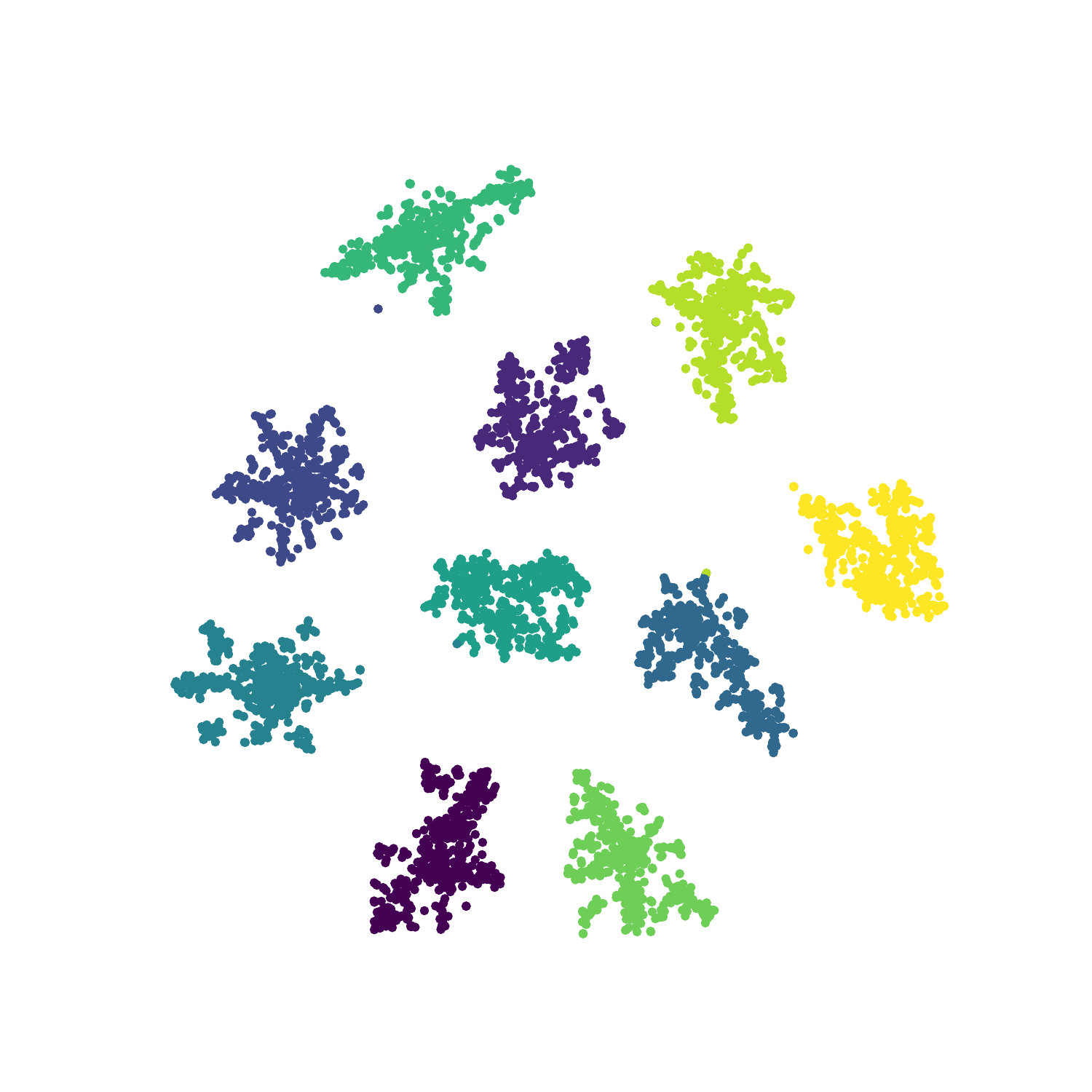}
}
\quad
\subfigure[\scriptsize{Epoch 250(ACC=0.995)}]{
\includegraphics[width=2.8cm]{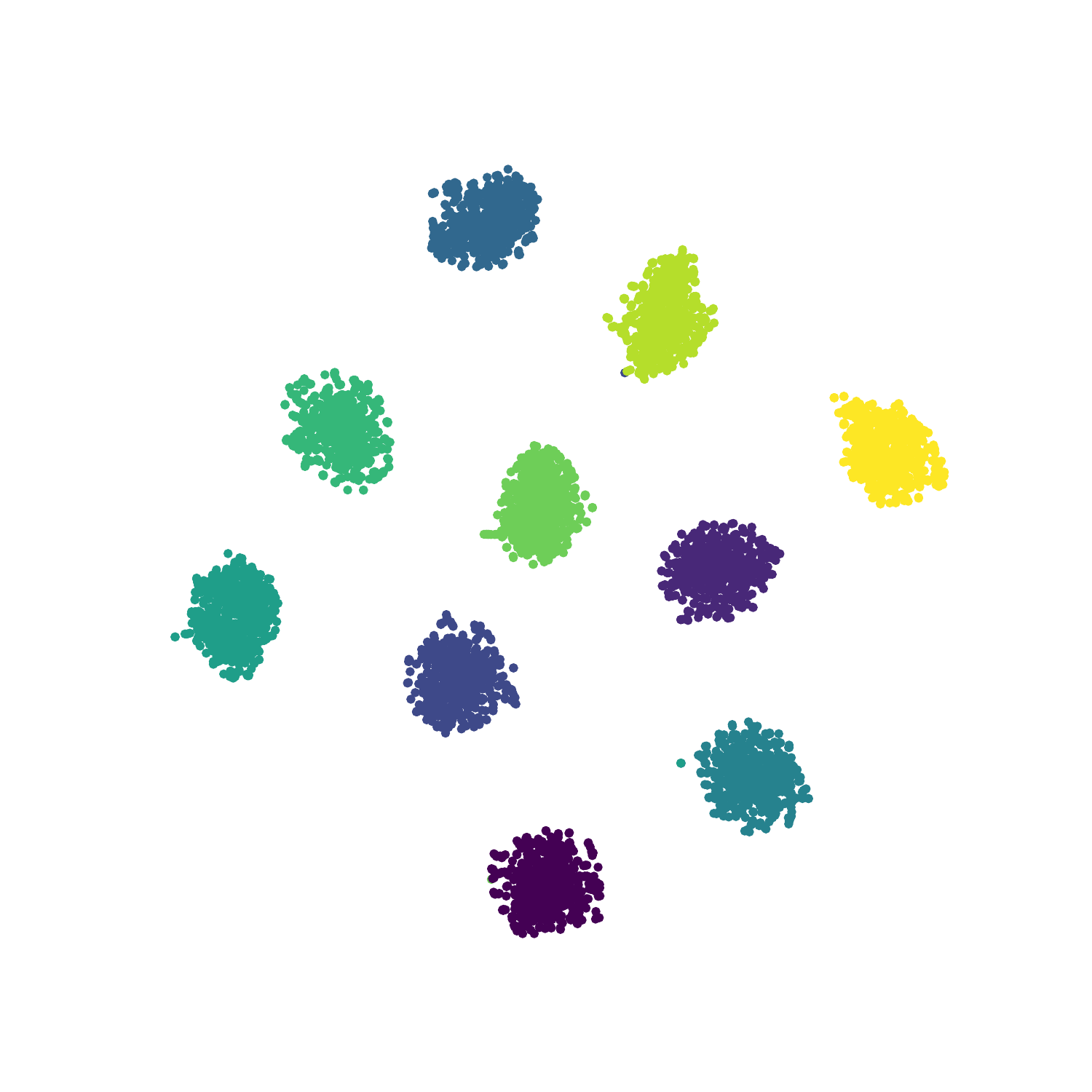}
}
\quad
\subfigure[\scriptsize{Epoch 10(ACC=0.290)}]{
\includegraphics[width=2.8cm]{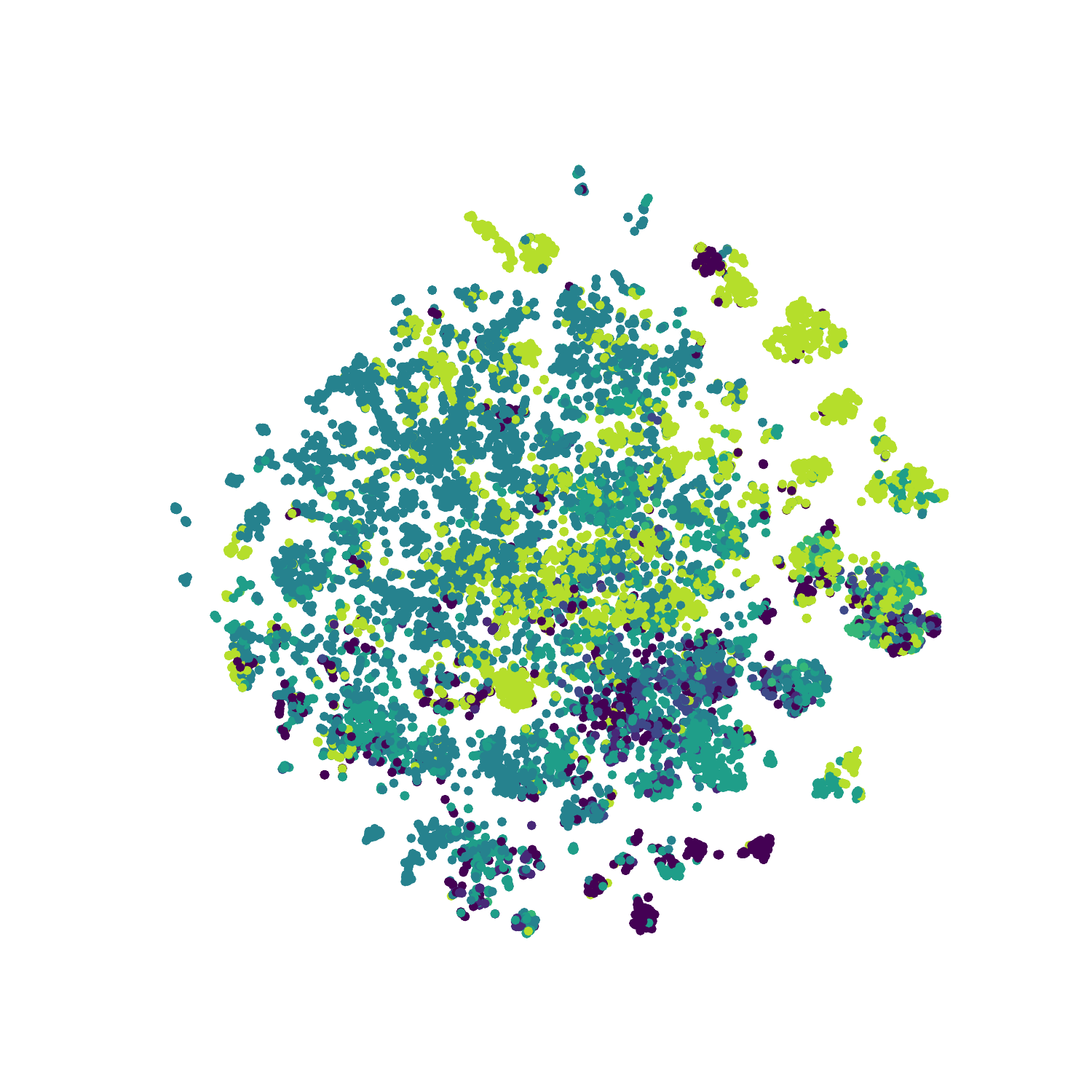}
}
\quad
\subfigure[\scriptsize{Epoch 50(ACC=0.286)}]{
\includegraphics[width=2.8cm]{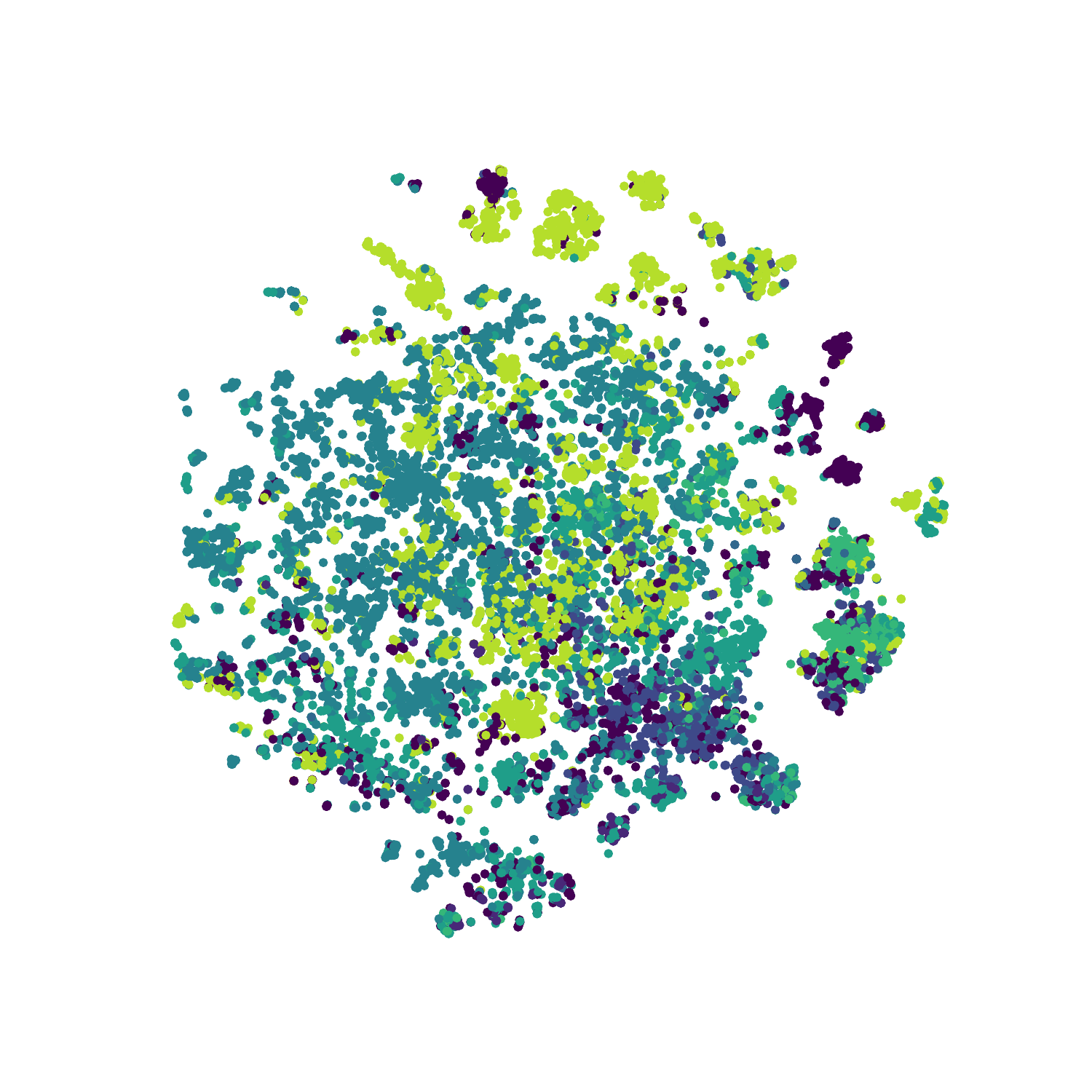}
}
\quad
\subfigure[\scriptsize{Epoch 100(ACC=0.371)}]{
\includegraphics[width=2.8cm]{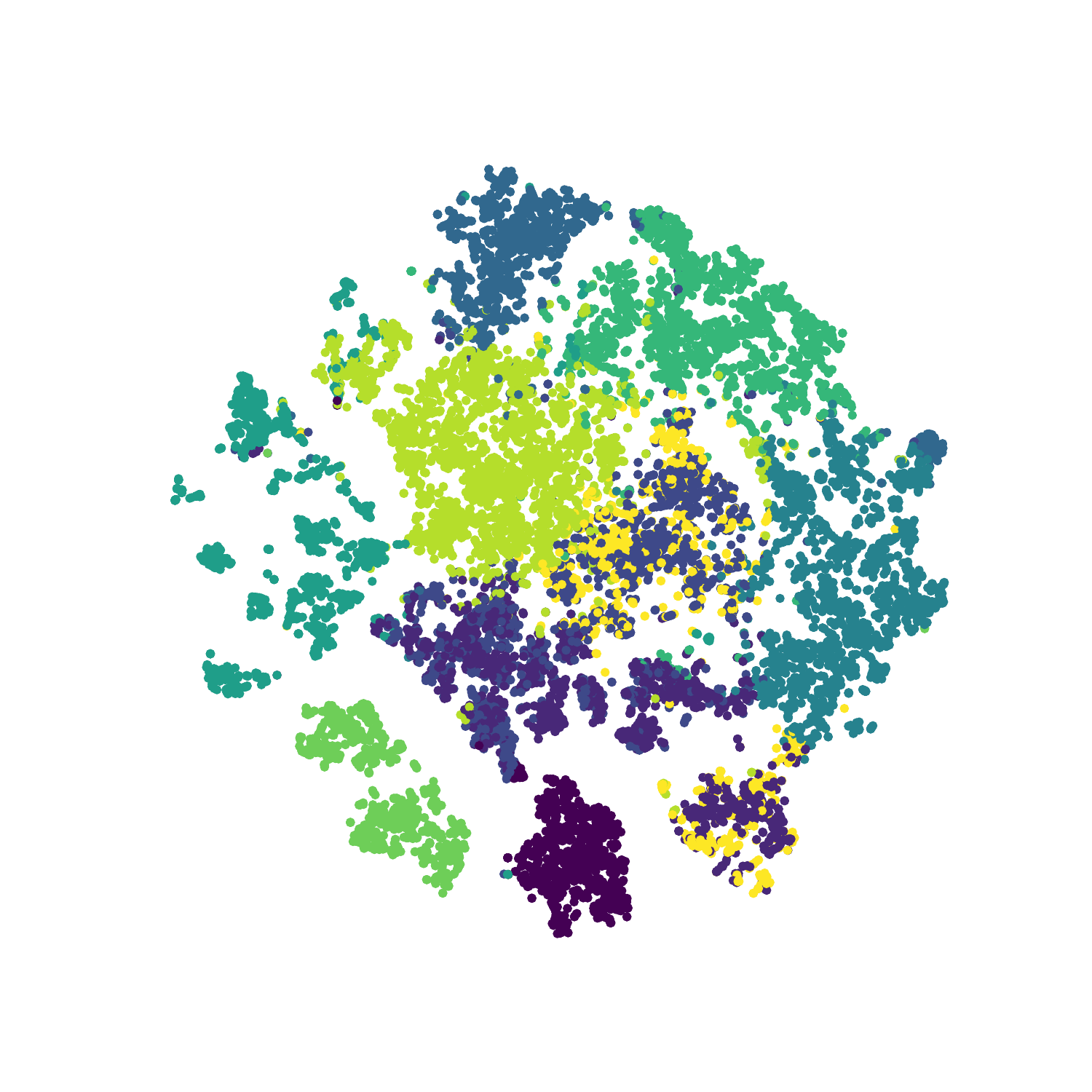}
}
\quad
\subfigure[\scriptsize{Epoch 200(ACC=0.446)}]{
\includegraphics[width=2.8cm]{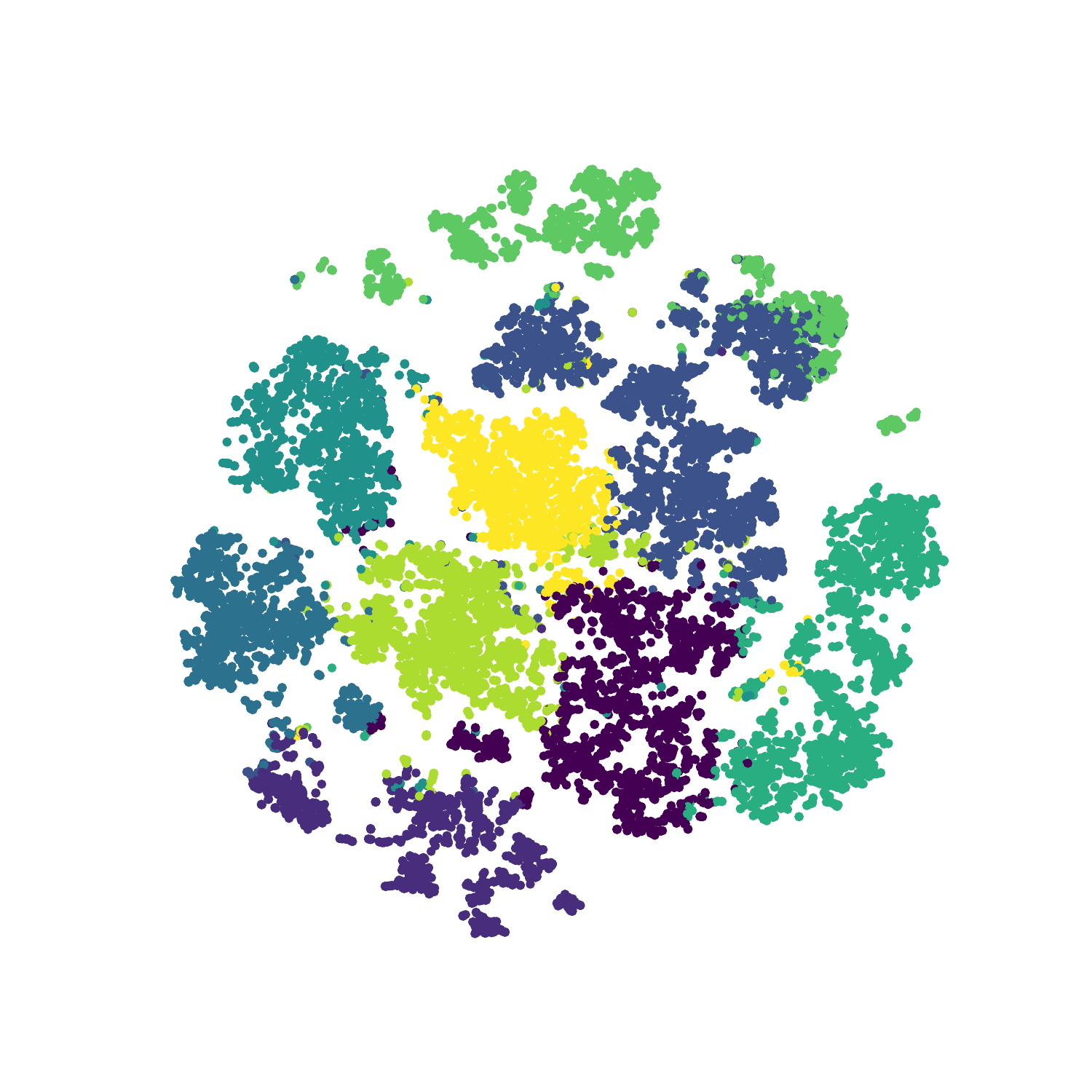}
}
\quad
\subfigure[\scriptsize{Epoch 250(ACC=0.465)}]{
\includegraphics[width=2.8cm]{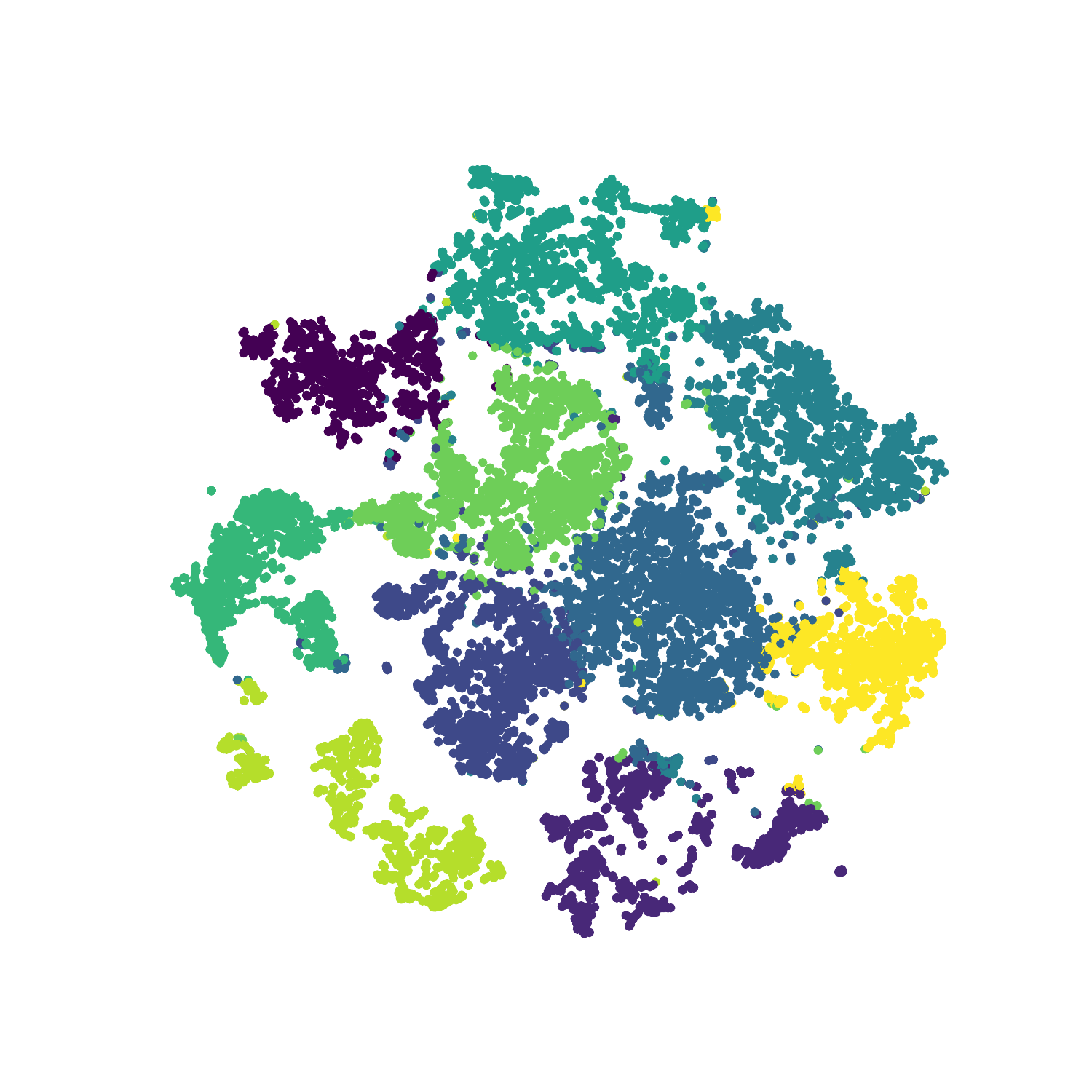}
}
\caption{
Visualization of common representations during training via t-SNE \cite{van2008visualizing}. In detail, the figure shows the visualization results of the representations of BBCSport (a-e), CCV (f-j), NMIST-USPS (k-o), and Reuters(p-t) when the epochs are 10, 50, 100, 200 and 250, respectively.}
\label{visualization}
\end{figure*}

\subsection{Ablation Study.}
\label{sec4.5}
We conduct experiments to verify the effectiveness of each components of the proposed method on BBCSport, CCV, MNIST-USPS, and Reuters datasets. For simplicity, we represent "R", "R\&S", "R\&S\&SI", "R\&S\&P", "R\&S\&P\&SI" and "R\&S\&P\&SI\&CL" as follows: (1) Reconstruction only, (2) Reconstruction and Shared information, (3) Reconstruction, Shared Information and Shared Information Inference, (4) Reconstruction, Shared Information and Private Information, (5) Reconstruction, Shared Information, Private Information and Shared Information Inference, (6) Reconstruction, Shared Information, Private Information, Shared Information Inference and cross-view Consistency Learning.From the results in Table. \ref{table3}, one can observe that. (i) In columns (2), (3) and (4), (5) of Table \ref{table4}, without the constraints of shared information, the consistency of shared information becomes weaker, leading to lower clustering performance. (ii) As far as (2) and (4) is concerned, the presence of private information significantly improves the clustering performance. This is because private information of a view cannot be expressed in common information, but it helps in view reconstruction. These result provides evidence of the ability of the proposed method to disentangle latent representations into shared information and private information. (iii) In columns (5) and (6), the introduction of contrastive learning reduces the effect of noise and redundancy among all views and maximizes the common information of the same sample cross-views. Since the consistency of the shared information and common representations is guaranteed, the learned representations are guaranteed, and private information is beneficial to reconstruction, thus achieving better performance. Overall, each component of the proposed method plays an integral role, and optimizing each component alone may lead to meaningless solutions.

\subsection{Visualization}
\label{sec4.6}
In this subsection, we conduct the t-SNE \cite{van2008visualizing} visualization experiment to demonstrate the superiority of the proposed method intuitively. To be specific, we visualized the distribution of the learned representations on BBCSport, CCV, MNIST-USPS, and Reuters. From the result in Fig. \ref{visualization}, it can be observed that the common representations learned become more compact and independent, and yields well-separated clusters with the epoch increases.


\section{CONCLUSION}
\label{sec5}
This paper proposes DCCMVC to provide a unifying framework for handing the challenging problem of the conflict between learning consistent common semantics and reconstructing consistency and complementarity information. It should be noted that the consistency and complementarity information in multi- view are not two separate parts. For capturing abstract relationships between views, disentangled and explainable visual representation learning can help learns more useful and generalized representations, which is instructive for multi-view learning. Therefore, we hope that future works can extend our framework to industrial-grade scenarios such as cross-modal feature retrieval, unsupervised labelling, and 3D reconstruction.

\printcredits

\bibliographystyle{cas-model2-names}

\bibliography{cas-refs}

\vskip3pt


\subsection*{  }
\setlength\intextsep{0pt}
\begin{wrapfigure}{l}{20mm}
    \centering
    \includegraphics[width=0.8in,height=1.25in,clip,keepaspectratio]{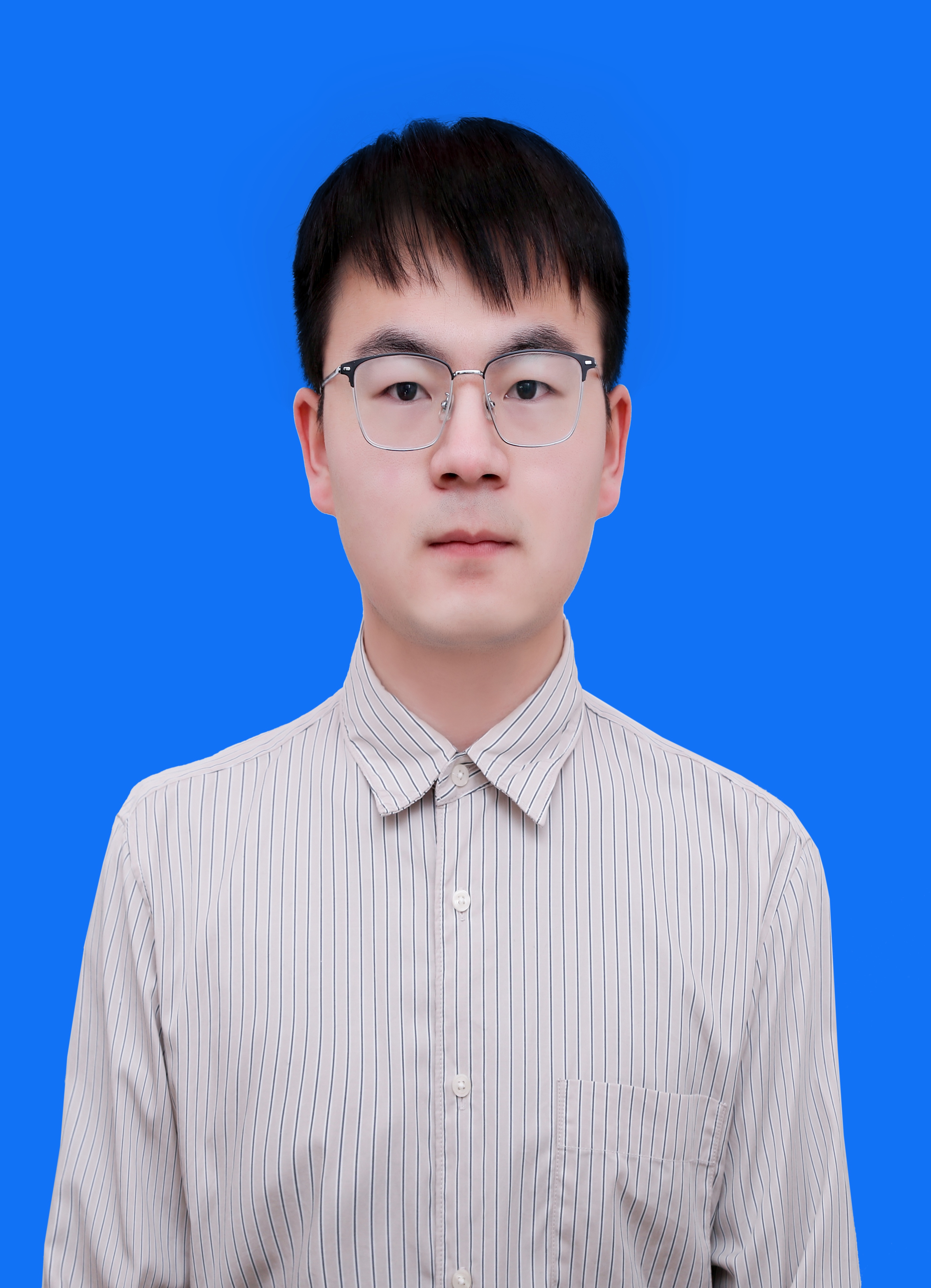}
\end{wrapfigure}
\noindent \textbf{Bo Li} received the M.S. degree from Inner Mongolia University of Technology, Hohhot, China. His research interests include pattern recognition and computer vision.\par

\hspace*{\fill} 

\subsection*{  }
\setlength\intextsep{0pt}
\begin{wrapfigure}{l}{20mm}
    \centering
    \includegraphics[width=0.8in,height=1.25in,clip,keepaspectratio]{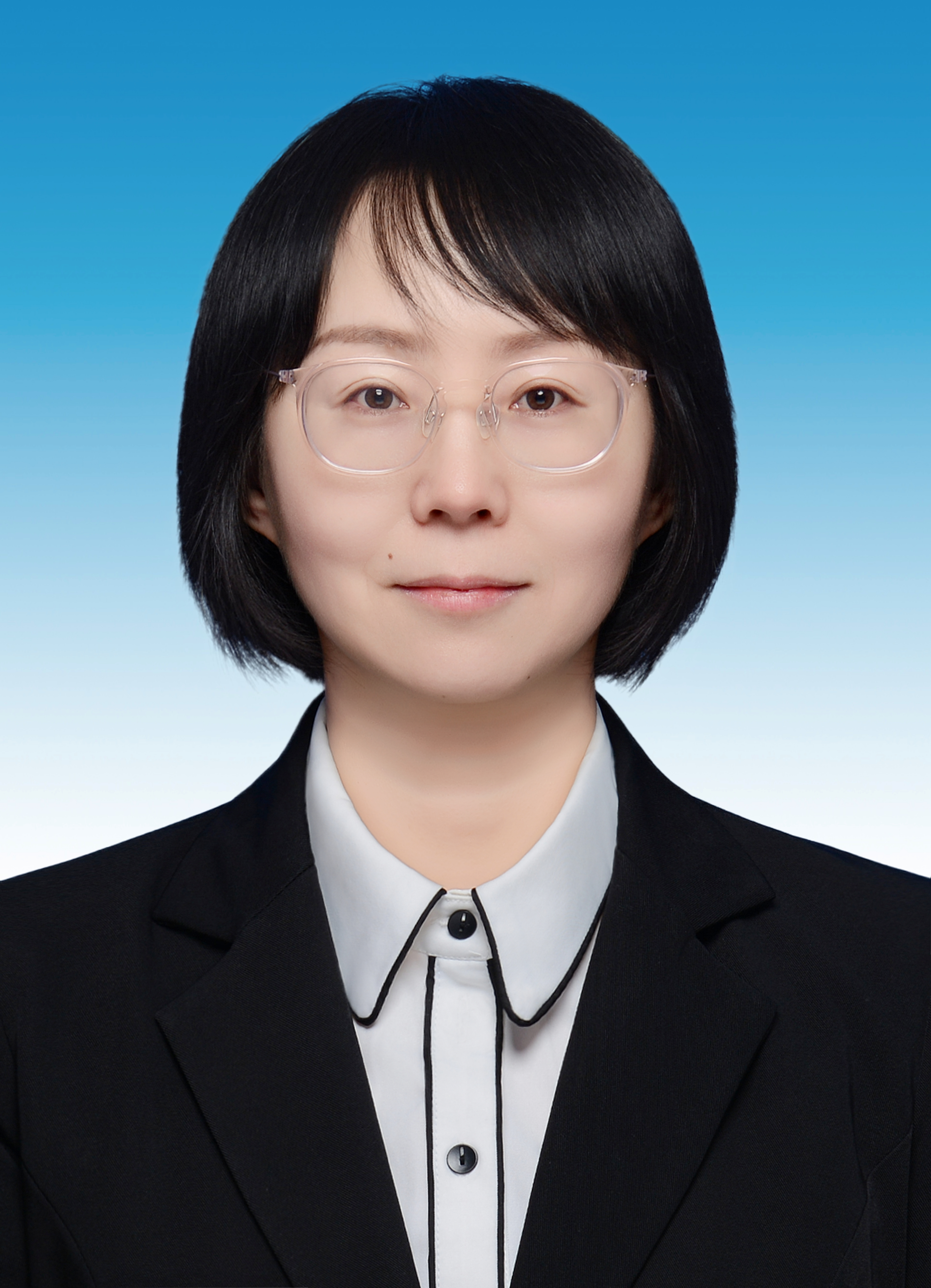}
\end{wrapfigure}
\noindent \textbf{Jing Yun} received the B.S. degree from Inner Mongolia University, Hohhot, China, in 2003, the M.S. degree from Beijing Information Science and Technology University, Beijing, China, in 2009, and the Ph.D. degree from Inner Mongolia University, in 2021. Her research interests include image processing, image caption, and machine learning algorithms. She is currently a Associate Professor with the College of Data Science and Application, Inner Mongolia University of Technology.


\end{document}